\documentclass[sigconf]{aamas} 

\usepackage{balance} 
\usepackage{amsmath}
\usepackage{mathtools}
\usepackage{amsthm}

\usepackage{verbatim}
\usepackage{cuted}
\usepackage{mathtools, bm}
\usepackage{latexsym,color}
\usepackage{algorithm}
\usepackage{algorithmic}
\usepackage{comment}
\usepackage{wrapfig}
\usepackage{makecell}
\usepackage{microtype}

\doi{UOYL1486}

\makeatletter
\gdef\@copyrightpermission{
  \begin{minipage}{0.2\columnwidth}
   \href{https://creativecommons.org/licenses/by/4.0/}{\includegraphics[width=0.90\textwidth]{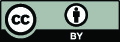}}
  \end{minipage}\hfill
  \begin{minipage}{0.8\columnwidth}
   \href{https://creativecommons.org/licenses/by/4.0/}{This work is licensed under a Creative Commons Attribution International 4.0 License.}
  \end{minipage}
  \vspace{5pt}
}
\makeatother

\setcopyright{ifaamas}
\acmConference[AAMAS '26]{Proc.\@ of the 25th International Conference
on Autonomous Agents and Multiagent Systems (AAMAS 2026)}{May 25 -- 29, 2026}
{Paphos, Cyprus}{C.~Amato, L.~Dennis, V.~Mascardi, J.~Thangarajah (eds.)}
\copyrightyear{2026}
\acmYear{2026}
\acmDOI{}
\acmPrice{}
\acmISBN{}

\title[AAMAS-2026 Formatting Instructions]{B3C: A Minimalist Approach to Offline Multi-Agent Reinforcement Learning}

\author{Woojun Kim}
\affiliation{
  \institution{Carnegie Mellon University}
  \city{Pittsburgh}
  \country{United States}}
\email{woojunk@andrew.cmu.edu}

\author{Katia P. Sycara}
\affiliation{
  \institution{Carnegie Mellon University}
  \city{Pittsburgh}
  \country{United States}}
\email{sycara@andrew.cmu.edu}

\begin{abstract}
Overestimation arising from selecting unseen actions during policy evaluation is a major challenge in offline reinforcement learning (RL). A minimalist approach in the single-agent setting---adding behavior cloning (BC) regularization to existing online RL algorithms---has been shown to be effective in terms of achieving competitive performance with minimal modification to online algorithms; however, this approach is understudied in multi-agent settings. In particular, overestimation becomes worse in multi-agent settings due to the presence of multiple actions, resulting in the BC regularization-based approach easily suffering from either over-regularization or critic divergence. To address this, we propose a simple yet effective method, Behavior Cloning regularization with Critic Clipping (B3C), which clips the target critic value in policy evaluation based on the maximum return in the dataset and pushes the limit of the weight on the RL objective over BC regularization, thereby demonstrating superior performance across benchmarks. Additionally, we leverage existing value factorization techniques, particularly non-linear factorization, which is understudied in offline settings. Integrated with non-linear value factorization, B3C outperforms state-of-the-art algorithms on various offline multi-agent benchmarks.
\end{abstract}

\keywords{Offline Multi-Agent Reinforcement Learning; Value Factorization}

\newcommand{\BibTeX}{\rm B\kern-.05em{\sc i\kern-.025em b}\kern-.08em\TeX}

\begin{document}


\pagestyle{fancy}
\fancyhead{}


\maketitle 


\section{Introduction}

Reinforcement learning (RL) trains agents to maximize cumulative rewards through interaction with an environment in an online manner; however, such interaction can be time-consuming due to the sample inefficiency of RL and costly in safety-critical environments due to inherent risks~\cite{levine2020offline, NEURIPS2023_a6f6a5c5}. Furthermore, in many practical scenarios, only pre-collected datasets can be leveraged instead of interacting with the environment. These challenges have motivated a shift toward offline RL, which trains agents using pre-collected datasets without requiring interaction~\cite{kostrikovoffline, wu2019behavior, agarwal2020optimistic, peng2023weighted}.

A major challenge of offline RL is that bootstrapping from unseen actions in policy evaluation induces extrapolation error, leading to accumulated errors~\cite{levine2020offline, fujimoto2021minimalist}, due to the inability to generate data. This issue consequently causes overestimation of value functions and adversely affects policy improvement. To address this, a variety of techniques, such as conservative value estimation~\cite{kumar2020conservative} and modeling behavior policy~\cite{fujimoto2019off}, have been introduced. However, these approaches increase algorithmic complexity, hinder reproducibility, and amplify hyperparameter sensitivity. This motivates minimalist approaches, which introduce minimal changes to existing RL algorithms~\cite{fujimoto2021minimalist, tarasov2024revisiting}. One representative example is TD3+BC, which integrates behavior cloning (BC) regularization into an existing online RL algorithm called TD3~\cite{fujimoto2021minimalist}. This simple approach does not require any additional components and introduces only one more hyperparameter for regularization, but it surprisingly achieves SOTA performance. This line of research, minimalist approaches, has been extensively investigated in single-agent settings~\cite{fujimoto2021minimalist, tarasov2024revisiting}, but it has barely been explored in multi-agent settings, which entail more algorithmic complexity due to the presence of multiple agents.

In this paper, we aim to develop a minimalist approach that introduces minimal changes to existing multi-agent RL algorithms. One might ask: is adding BC regularization sufficient in multi-agent settings? We argue that BC regularization alone is not enough in multi-agent settings. This is because the overestimation becomes more severe in multi-agent settings. It arises from the use of a centralized critic, which conditions on multiple actions, increasing the likelihood of encountering unseen actions. Consequently, critic learning often becomes unstable, necessitating a greater reliance on BC regularization over the RL objective---resulting in over-regularization. This over-regularization inherently limits performance, as it becomes heavily dependent on the quality of the dataset. In addition, recent research on offline multi-agent RL overlooks existing value factorization techniques, such as non-linear decomposition---they either rely on linear factorization or disregard factorization entirely.

To address the aforementioned limitations, we propose a simple yet effective technique for offline multi-agent RL: \textbf{B}ehavior \textbf{C}loning regularization with \textbf{C}ritic \textbf{C}lipping (\textbf{B3C}). The proposed method alleviates overestimation by clipping the critic value during policy evaluation, using the maximum return from the given dataset, in addition to BC regularization. Critic clipping enables the use of a higher weight for the RL objective relative to BC regularization, resulting in improved performance. Furthermore, we integrate B3C with an existing online multi-agent RL algorithm called FACMAC~\cite{peng2021facmac}, which utilizes factored critics and value factorization techniques, resulting in \textbf{FACMAC+B3C}. Notably, we provide a design choice for value factorization in the offline setting, empirically demonstrating that non-monotonic factorization performs better than monotonic and linear factorization, which contrasts with findings in the online setting~\cite{peng2021facmac}.

Despite its simplicity, the effectiveness of FACMAC+B3C is empirically demonstrated across a variety of offline multi-agent RL benchmarks, including multi-agent Mujoco and particle environments, with various types of dataset. We provide a thorough analysis demonstrating how the proposed method outperforms the baselines in terms of performance. Our main contributions are:\\
$\bullet$ We address the over-regularization problem in offline multi-agent RL and propose a minimalist framework that achieves  SOTA performance with minimal modification to existing algorithms.\\
$\bullet$ We conduct an empirical study of value factorization in offline settings, an aspect that has been understudied.\\
$\bullet$ We conduct extensive experiments across seven environments and forty-two datasets, covering both fully and partially observable settings.

\section{Background and Related Works}

\subsection{Multi-Agent RL}\label{subsec:multiagentrl}

We consider fully cooperative multi-agent tasks, which can be modeled as an N-agent decentralized partially observable MDP (Dec-POMDP)~\cite{oliehoek2012decentralized, dibangoye2018learning, amato2013decentralized}. At each timestep $t$, each agent selects its own action, $a_t^i$, based on its local information such as the partial observation $o_t^i$. This forms a joint action $\bm{a}_t$ yielding the next state $s_{t+1}$, the joint observations $\bm{o}_{t+1}=(o^1_{t+1},\cdots, o^N_{t+1})$, and a shared reward $r_t$. The goal is to find the optimal joint policy that maximizes the team return, $\mathbb{E}[\sum_{l=0}^{\infty}\gamma^l r_l]$. For this, recent multi-agent RL algorithms adopt a framework of centralized training with decentralized execution (CTDE)~\cite{wang2020dop, zhang2021fop, jeon2022maser, kim2023variational}, where individual policies execute actions based on local information, e.g., Agent $i$'s action-observation history $\tau^i$,  but are trained with global information such as the state.

One of the key challenges in Dec-POMDPs is to correctly assign credit to each agent for the team reward. For this, value factorization methods have been actively studied~\cite{sunehag2017value, rashid2020monotonic, son2019qtran, zhang2021fop, wang2021towards, marchesini2025stateful}. Value factorization methods approximate the joint $Q$-function as a function of individual agent-wise value functions, thereby enabling effective credit assignment among agents. Linear factorization, such as Value Decomposition Networks (VDN)~\cite{sunehag2017value}, assumes additive decomposition of individual $Q^i$ values, while monotonic factorization in QMIX~\cite{rashid2020monotonic}, enforces a monotonic mixing constraint to ensure consistency between local and global optima. Such methods have shown promising results in online multi-agent RL, but their applicability to offline settings remains unexplored.

\textbf{FACMAC}~\cite{peng2021facmac} is a representative multi-agent RL algorithm that trains a centralized but factored critic, which is decomposed into a nonlinear combination of individual critics, and uses it to train decentralized deterministic policies.
The key idea is factorizing the centralized critic into individual critics via a non-linear function, where the centralized critic is decomposed as 
\begin{align}
    Q_{jt}(s, \bm{\tau}, \textbf{a}) = f_{mixer}(s, Q^1(\tau^1, a^1), \cdots, Q^N(\tau^N, a^N)), 
\end{align} 
where $f_{mixer}$ is a mixer network that encodes state information and learns the weights of individual critics, and $Q^i(\tau^i, a^i)$ is the individual critic of Agent $i$. FACMAC considers two value factorization methods: (a) monotonic factorization (\textit{mono}), which constrains the parameters of $f_{mixer}$ to enforce $\partial Q_{jt}/ \partial Q^i \geq 0$, as in QMIX~\cite{rashid2020monotonic}; and (b) non-monotonic factorization (\textit{non-mono}), which removes the constraint from (a) to increase representational capacity. In addition to these two methods (c) linear factorization~\cite{sunehag2017value} (\textit{vdn}) was also considered. It was shown that \textit{non-mono} and \textit{mono} approaches are complementary: \textit{non-mono} performs better on tasks requiring greater representational capacity, while \textit{mono} outperforms on others. However, \textit{vdn} underperforms both and performs drastically worse, particularly in multi-agent Mujoco environment.

The centralized but factored critic, parameterized by $\bm{\theta_{Q}}$, is trained to minimize the temporal-difference (TD) error:
\begin{align}
    \mathcal{L}_{\text{FACMAC}}&(\bm{\theta_{Q}}) = \mathbb{E}_{\mathcal{D}}\left[ \Big(y^{jt} -Q_{jt}(s, \bm{\tau}, \bm{a};\bm{\theta_{Q}})\Big)^2\right], \nonumber ~ \mbox{where} \nonumber \\ &y^{jt}=r+\gamma Q_{jt}(s', \bm{\tau'}, \bm{\pi}(\bm{\tau'});\bm{\theta_{Q}}^-),
\end{align}
$\mathcal{D}$ is the replay buffer, $\bm{\pi}$ is the deterministic joint policy, and $\bm{\theta_{Q}^-}$ is the parameter of the target critic networks. Based on this centralized critic, the decentralized policies, parameterized by $\bm{\theta}_{\bm{\pi}}=\{\theta_{\pi}^i\}_{i=1}^N$, are trained using the deterministic policy gradient~\cite{silver2014deterministic} where the loss function is written as $\mathcal{L}_{\text{FACMAC}}(\bm{\theta}_{\bm{\pi}}) =$
\begin{align}\label{eq:facmac_actor}
    \mathbb{E}_{\mathcal{D}}\left[ -Q_{jt}(s, \bm{\tau}, \pi^1(\tau^1;\theta_{\pi}^1), \cdots, \pi^N(\tau^N;\theta_{\pi}^N))\right], 
\end{align}
where $\theta_{\pi}^i$ is Agent $i$'s actor parameter. A main benefit of FACMAC over prior works such as MADDPG~\cite{lowe2017multi} is leveraging non-linear factorization. However, in offline settings, value factorization techniques remain underexplored---recent research still relies on linear factorization~\cite{yang2021believe, wang2024offline} or omits it by using a non-factored critic. In this paper, to investigate value factorization for offline settings, we adopt FACMAC to build upon the recent achievements.

\subsection{Offline RL}\label{sec:offlineRL}

Offline RL aims to learn a policy that maximizes the expected return from a fixed dataset $\mathcal{D}$, consisting of trajectories generated by arbitrary behavior policies, without additional environment interactions~\cite{kumar2020conservative, kostrikov2022offline, kim2024adaptive}. This inability to generate additional data introduces a major challenge in offline RL---\textit{extrapolation error} in policy evaluation, i.e., the target value in the Bellman equation becomes inaccurate if actions from the learning policy are not included in the dataset. This extrapolation error often leads to overestimation and degrades performance. To address this challenge, several approaches have been proposed, such as behavior-constrained policy optimization, which regularizes the policy to stay close to the behavior policy~\cite{wu2019behavior, fujimoto2021minimalist, wu2022supported, zhang2023constrained}, and conservative value estimation~\cite{kumar2020conservative, kostrikov2021offline}, which lower-bounds the value function by penalizing the values of unseen actions. These approaches have demonstrated effectiveness; however, they often require significant modifications or additional components to existing RL algorithms, such as generative models. These additions introduce more hyperparameters, which can affect performance and impact stability and reproducibility. This necessitates an offline RL algorithm with minimal changes from the current online RL algorithms. As an example, \cite{fujimoto2021minimalist} proposed TD3 with Behavior Cloning (TD3+BC), which adds a BC loss function as a regularizer on top of the TD3 loss~\cite{fujimoto2018addressing} to update the policy. This encourages the action generated by the learning policy to be the same as the action in the dataset. The policy of TD3+BC is updated to minimize the following loss function.
\begin{align}\label{eq:td3+bc}
    \mathcal{L}_{\text{TD3+BC}}(\pi) =  \mathbb{E}_{\mathcal{D}}&\Big[-\bm{w} \underbrace{Q(s,\pi(s))}_{TD3} + \underbrace{(\pi(s)-a)^2}_{BC} \Big],  \nonumber \\ \mbox{where} ~~~~~~~~~~&\bm{w} = \frac{\alpha}{\frac{1}{N} \sum|Q(s,\pi(s))|}
\end{align}
where $\alpha$ is a hyperparameter that controls the balance between RL and BC. TD3+BC has been widely used due to its simplicity and performance comparable to SOTA algorithms~\cite{kim2024decision}.

\subsection{Offline Multi-Agent RL}\label{sec:offline_marl}

A natural approach to offline multi-agent RL is to extend the previously mentioned single-agent offline RL to the multi-agent setting. However, naively extending these methods has been shown to be insufficient~\cite{pan2022plan, shao2024counterfactual}. For example, naive conservative value estimation for the joint policy can lead to excessively large penalties, significantly degrading performance as the number of agents increases~\cite{shao2024counterfactual}. Furthermore, the problem of extrapolation error worsens as the size of the action space grows with the number of agents~\cite{yang2021believe}, making it essential to balance conservatism and regularization with extrapolation error. To mitigate this problem, CFCQL~\cite{shao2024counterfactual} introduces counterfactual regularization to each agent individually to alleviate excessive conservatism. OMAR~\cite{pan2022plan} introduces zeroth-order optimization to address policies getting stuck in poor local optima under conservative value function training. Note that both OMAR and CFCQL are based on conservative value estimation, which penalizes value functions for unseen action spaces, thereby forcing them to underestimate. OMIGA~\cite{wang2024offline} introduces a framework that converts global regularization into implicit local-level regularization using linear value decomposition, while employing in-sampling techniques to avoid querying unseen actions. MADIFF~\cite{zhu2023madiff} presents a diffusion-based generative approach for multi-agent systems, enabling decentralized policy development through effective teammate modeling.

Although the aforementioned offline multi-agent RL algorithms perform reasonably, they still require significant modifications from existing online multi-agent RL algorithms, as described in single-agent settings in Sec. \ref{sec:offlineRL}, with even more adjustments needed due to the presence of multiple agents. Additionally, they have not fully leveraged the currently existing value factorization techniques, which are central to the recent success of multi-agent RL. Therefore, inspired by \cite{fujimoto2021minimalist}, we aim to develop a minimalist approach that minimally modifies an existing multi-agent RL algorithm, focusing on simplicity and practicality while incorporating existing multi-agent RL techniques.

\section{Methodology}

To achieve the aforementioned goal, we focus on a BC regularization-based approach. In multi-agent settings, the presence of multiple agents amplifies the issue of overestimation, as mentioned in Sec.\ref{sec:offline_marl}. Consequently, the BC regularization-based approach is susceptible to either over-regularization or unstable critic learning. This occurs because the joint action space grows exponentially with the number of agents, increasing the likelihood of encountering unseen actions during policy evaluation. As a result, critic updates become unstable and prone to divergence, which in turn encourages excessive reliance on the BC term to stabilize training, leading to over-regularization and performance degradation when the dataset quality is limited. 

To address these challenges, we propose a simple yet effective regularization method called \textbf{B}ehavior \textbf{C}loning with \textbf{C}ritic \textbf{C}lipping (\textbf{B3C}), which prevents over-regularization while ensuring stability, allowing a focus on the RL objective and thereby significantly improving performance. Additionally, we explore the application of value factorization techniques with B3C, which remains understudied.

\begin{figure}[t]
\centering
\includegraphics[width=0.9\linewidth]{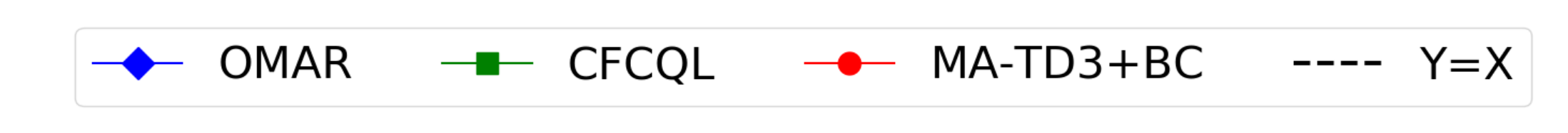}
\begin{tabular}{cc}
\hspace{-2ex}\includegraphics[width=0.5\linewidth]{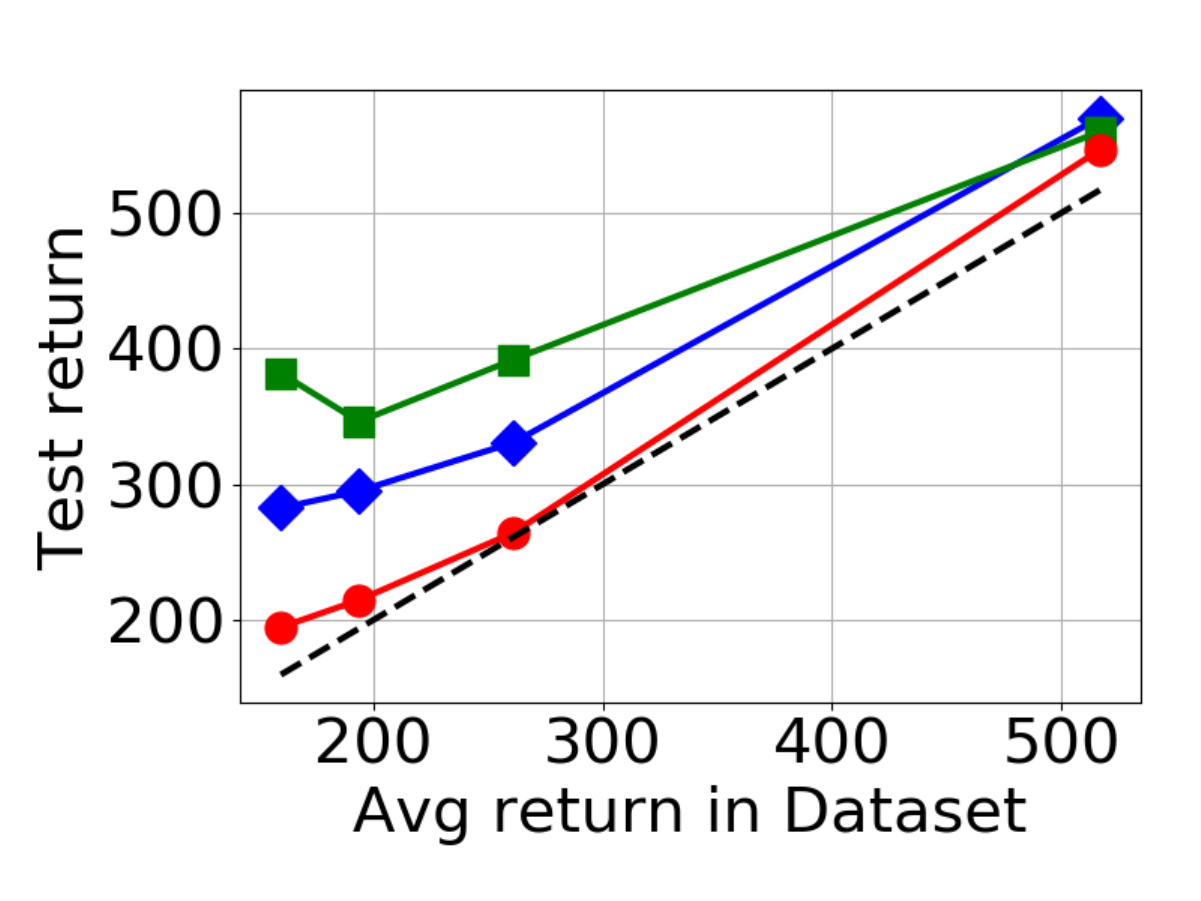} \hspace{-3ex} &
\includegraphics[width=0.5\linewidth]{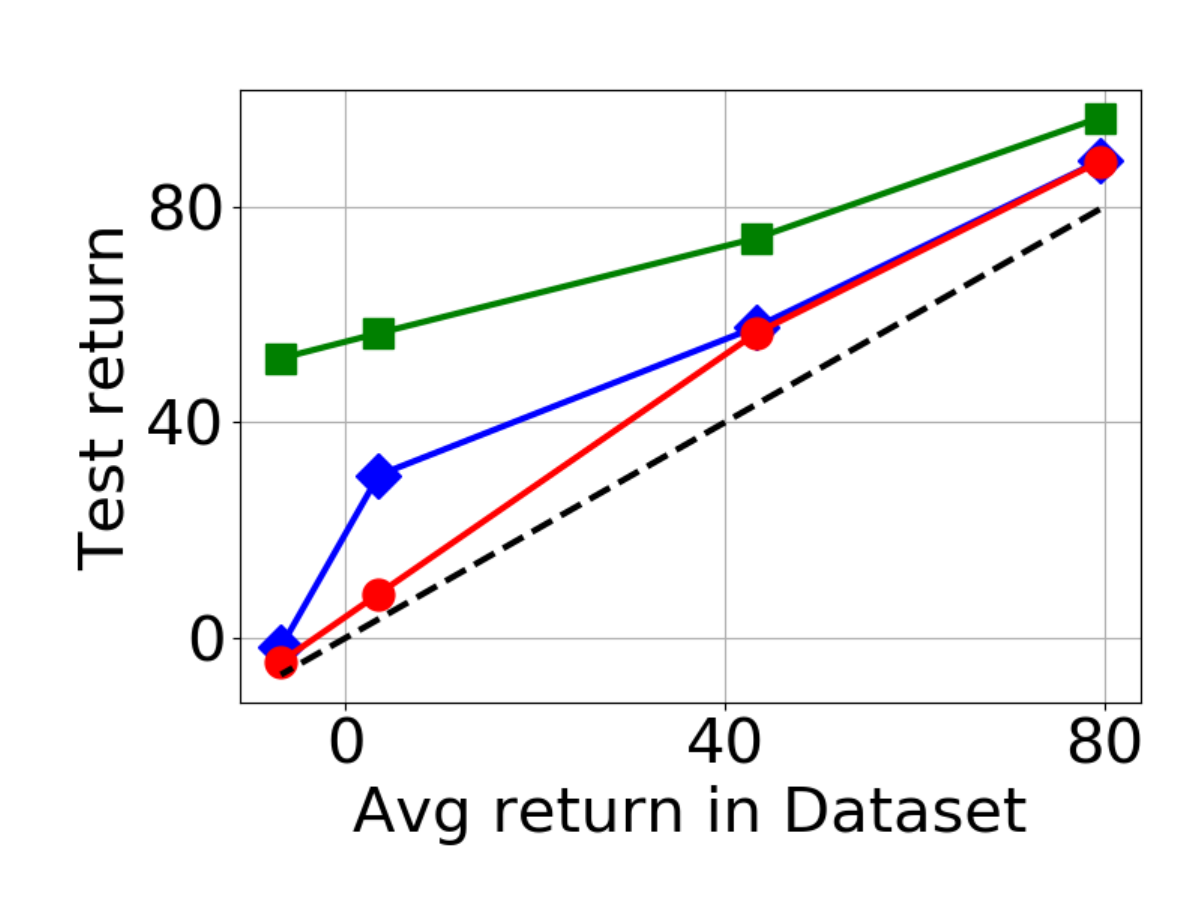} \\
\hspace{7ex}(a) CN  & \hspace{3ex} (b) World
\end{tabular}

\caption{Test return as a function of dataset quality (average dataset return) in CN and World, based on \cite{shao2024counterfactual}. Proximity to $y = x$ reflects strong data dependence. MA-TD3+BC remains close to dataset-level performance, indicating over-regularization by behavior cloning.
}
\vspace{-2ex}
\label{fig:performance_td3}
\end{figure}

\subsection{Revisiting Behavior Cloning: Over-regularization}\label{sec:overreg}

We begin by revisiting BC regularization in offline multi-agent RL. Building on the success of BC regularization in the single-agent setting, 
recent studies on offline multi-agent RL~\cite{pan2022plan, shao2024counterfactual} have introduced an extension of TD3+BC, called MA-TD3+BC, as a baseline by adapting Eq. \ref{eq:td3+bc} to the multi-agent setting. Although MA-TD3+BC achieves reasonable performance, it is less effective than the proposed methods, particularly when applied to low-quality datasets. Fig. \ref{fig:performance_td3} illustrates the test returns of OMAR~\cite{pan2022plan}, CFCQL~\cite{shao2024counterfactual}, and MA-TD3+BC with respect to the average return of the dataset, reported in \cite{shao2024counterfactual}. MA-TD3+BC achieves performance close to dataset-level quality, but with a small margin. In contrast, OMAR and CFCQL show significant improvements over dataset quality, particularly for low-quality datasets. This observation, that MA-TD3+BC heavily depends on data quality, indicates that it is over-regularized by behavior cloning.

One might ask: what if we place more weight on the RL objective in Eq. \ref{eq:td3+bc}, which can be achieved by increasing $\alpha$ in Eq. \ref{eq:td3+bc}, to reduce dependence on data quality? We observed that increasing $\alpha$ sometimes leads to better performance compared to prior work; however, it often fails to converge. The critic is updated unstably and even frequently diverges due to extrapolation error, as we will discuss in Sec.~\ref{sec:b3c_versus_bc}. Therefore, to achieve stable performance, $\alpha$ must be kept low, which results in over-regularization and limits performance based on dataset quality.

\subsection{Critic Clipping: Alleviate Overestimation}\label{sec:cc}

In order to simultaneously avoid unstable critic training and over-regularization, we propose a simple yet effective technique called Critic Clipping (\texttt{CC}) for training the centralized critic, which integrates with BC regularization for training policies.

\texttt{CC} clips the critic target value when calculating the target value during policy evaluation, which can lead to overestimation based on the maximum return in the dataset. Unlike conservative value estimation, which requires additional sampling or modeling of the behavior policy and introduces underestimation bias, \texttt{CC} is easy to implement and proves effective by limiting overestimation to below a certain threshold. Intuitively, this prevents the critic from propagating excessively optimistic targets that would otherwise destabilize value learning across agents. Here, the reason for using the maximum return rather than the average return as the clipping standard is as follows: The learned RL policy is expected to outperform the behavior policy, and therefore, the expected return, which the critic approximates, is expected to be larger than the average return in the dataset. For simplicity, we use the maximum return, representing the best episode in the dataset. To allow further flexibility, we can use a scaled value of the maximum return for the clipping value. Note that we observe using the maximum return is sufficient empirically, as we will discuss in Sec.\ref{sec:clipping_value}. The corresponding loss function of the critic is given by 
\begin{align}\label{eq:loss_cc}
    &\mathcal{L}_{CC}(\bm{\phi}, \psi)= \mathbb{E}_{(s,\bm{\tau}, \bm{a}, r, s', \bm{\tau'})\sim \mathcal{D}}\left[ \Big( y^{jt} - Q_{jt}(s, \bm{\tau}, \bm{a}; \bm{\phi}) \Big)^2\right], ~~ \mbox{where} \nonumber \\ &y^{jt} = r+\gamma ~ \underbrace{\mbox{Min}\Big[Q_{jt}(s', \bm{\tau'}, \bm{\pi}(\bm{\tau'};\bm{\theta^{-}}); \bm{\phi^{-}}), ~~ R^*\Big]}_{\text{Critic Clipping}} ~ \mbox{and}
\end{align}
$R^*=M\times \max_{d}\sum_{t=1}^T r_{d, t}$. Here, $M$ is a scalar value, and $\max_{d}\sum_{t=1}^T r_{d, t}$ represents the maximum return in the dataset where $r_{d,t}$ is the reward at time-step $t$ of $d$-th episode in the dataset. For simplicity, we use $M=1$ in most cases during the experiments. 
We observed that this straightforward choice is sufficient in most cases, aligning with our minimalist design philosophy and reducing the burden of hyperparameter tuning. Although $M$ can be fine-tuned (e.g., smaller $M$ performed slightly better in one environment; see Appendix~\ref{appendix:hyper}), we did not observe a clear or consistent improvement from adjusting it. Additionally, this clipping does not constrain the learned policy to achieve returns below the maximum return in the dataset. The critic value provides an approximation of policy performance but does not constitute a hard upper bound, as its estimates are used for optimization guidance rather than to determine the final achievable return. We provide an ablation study on this in Sec. \ref{sec:clipping_value}.

\subsection{B3C with Value Factorization}\label{sec:b3c_vf}

We integrate \texttt{CC} with BC regularization, introducing the proposed method, B3C. The main advantage of \texttt{CC} is its flexibility, which reduces sensitivity to BC regularization and allows for maximizing the weight of the RL objective. This ultimately leads to improved performance. Additionally, we adopt FACMAC, which uses a centralized but factored critic with non-linear value factorization. Accordingly, the policy loss function of FACMAC+B3C is given by $\mathcal{L}_{\text{FACMAC+B3C}}(\bm{\theta}_{\bm{\pi}}) = $
\begin{align}\label{eq:facmac+b3c_policyupdate}
     \mathbb{E}_{\mathcal{D}}\Big[-\bm{w}& \underbrace{Q_{jt}(s, \bm{\tau}, \pi^1(\tau^1), \cdots, \pi^N(\tau^N))}_{FACMAC} 
 + \beta \sum_{i=1}^N \underbrace{(\pi^i(\tau^i)-a^i)^2}_{BC} \Big] \nonumber \\ & \mbox{where} ~ \bm{w}= \frac{\alpha}{\frac{1}{N} \sum |Q_{jt}(s, \bm{\tau}, \bm{a})|},
\end{align}
$\bm{\pi} = (\pi^1, \cdots, \pi^N)$ and $\bm{a} = (a^1, \cdots, a^N)$ represent the joint policy and the joint action, respectively. Here, $\bm{\theta}_{\bm{\pi}}$ denotes the parameters of the joint policy, and $\mathcal{D}$ represents the given offline dataset. We normalize the Q-value based on the sample batch to ensure robustness to scale, following \cite{fujimoto2021minimalist}. Note that there are two important hyperparameters, $\alpha$ and $\beta$, which balance the trade-off between RL and BC. We will refer to $\alpha$ and $\beta$ as the RL coefficient and the BC coefficient, respectively. A key difference from \cite{fujimoto2021minimalist} is the introduction of the BC coefficient, $\beta$, alongside the RL coefficient, $\alpha$. This decouples the relative RL-BC weighting and the overall objective scale, enabling independent control of regularization strength, which can affect optimization in deep networks. For hyperparameter tuning, we recommend practitioners start by fixing $\beta = 1$ and tuning $\alpha$, following \cite{fujimoto2021minimalist}. If over-regularization persists, $\beta$ can then be adjusted, which has been shown to be effective in multi-agent particle environments.

\textbf{Empirical Observation regarding value factorization:} We consider three value factorization methods discussed in Sec.\ref{subsec:multiagentrl}: \textit{non-mono}, \textit{mono}, and \textit{vdn}. We empirically observe that \textit{non-mono} outperforms \textit{mono} in most cases, which differs from the findings in the online setting discussed in Sec. \ref{subsec:multiagentrl}. We argue that limiting representational capacity, which is inherent to monotonic factorization, leads to reduced performance in offline settings. We provide an ablation study for this in Sec. \ref{sec:ana_vf}.

As summarized above, we update the joint policy to maximize the critic with BC regularization by minimizing Eq. \ref{eq:facmac+b3c_policyupdate}, and we update the centralized critic to minimize the TD error with clipping, as shown in Eq. \ref{eq:loss_cc}. We refer to the proposed algorithm---FACMAC integrated with Behavior Cloning regularization and Critic Clipping---as FACMAC+B3C. Additionally, we also apply B3C to MA-TD3, referred to as MA-TD3+B3C.

\newcolumntype{P}[1]{>{\centering\arraybackslash}p{#1}}

\begin{table*}[t!]
\footnotesize
\caption{Performance of MA-TD3+B3C and baselines on multi-agent particle environments, averaged over five seeds. Tasks include Cooperative Navigation (CN), Predator-Prey (PP), and World, each evaluated on expert (e), medium (m), medium-replay (mr), and random (r) datasets. “Avg R” and “Max R” denote the dataset average and maximum return. Boldface indicates the best or statistically comparable performance.
}
\vspace{-2ex}
\centering
\renewcommand{\arraystretch}{1.2}
\begin{tabular}{|p{2cm}P{1cm}P{1cm}||P{1.2cm}P{1.2cm}P{1.2cm}P{1.2cm}|P{1.9cm}P{2.1cm}|}
 \hline
 \textbf{Task-Dataset}& Avg R & Max R &   \textbf{OMAR} & \textbf{CFCQL} &  \textbf{MADIFF}  &
 \textbf{TD3+BC} &
 \textbf{TD3+BC (ours)} & \textbf{TD3+B3C (ours)} \\
 \hline
 CN-e & 100.0 & 163.1 & \bf{114.9$\pm$2.6} & 112.0$\pm$4.0  & 95.0$\pm$5.3 & 108.3$\pm$3.3 &  100.6$\pm$8.3 & 102.3$\pm$6.7\\
 CN-mr & 9.5 & 123.2 & 37.9$\pm$12.3 & \bf{52.2$\pm$9.6}  & 30.3$\pm$2.5  & 15.4$\pm$5.6 &  \bf{53.8$\pm$4.8} & \bf{53.8$\pm$4.8}\\
 CN-m & 27.8 & 145.0 & 47.9$\pm$18.9 & \bf{65.0$\pm$10.2}  & 64.9$\pm$7.7  & 29.3$\pm$4.8 &  55.9$\pm$25.8 & \bf{63.5$\pm$9.7}\\
 CN-r & 0.0 & 103.7 & 6.9$\pm$3.1 & 62.2$\pm$8.1  & 6.9$\pm$3.1  & 9.8$\pm$4.9 &  \bf{72.6$\pm$7.0} & \bf{73.3$\pm$6.6}\\
 \hline
 PP-e & 100.0 & 275.7 & 116.2$\pm$19.8 & 118.2$\pm$13.1  & 120.9$\pm$14.6  & 115.2$\pm$12.5 &  108.1$\pm$22.9 & \bf{124.6$\pm$21.2}\\
 PP-mr & 9.6 & 128.0 & 47.1$\pm$15.3 & 71.1$\pm$6.0  &  62.3$\pm$9.2  & 28.7$\pm$20.9 &  \bf{75.2$\pm$6.8} & \bf{76.5$\pm$8.6}\\
 PP-m & 48.9 & 196.2 & 66.7$\pm$23.2 & 68.5$\pm$21.8  & 77.2$\pm$10.4   & 65.1$\pm$29.5 &  91.3$\pm$35.7 & \bf{104.3$\pm$11.9}\\
 PP-r & 0.0 & 61.4 & 11.1$\pm$2.8 & 78.5$\pm$15.6  & 3.2$\pm$4.0  & 5.7$\pm$3.5 &  \bf{105.0$\pm$16.7} & \bf{105.5$\pm$10.0}\\
 \hline
 World-e & 100.0 & 270.7 & 110.4$\pm$25.7 & 119.7$\pm$26.4  & \bf{122.6$\pm$14.4}  & 110.3$\pm$21.3 &  120.7$\pm$17.9 & \bf{124.3$\pm$5.1}\\
 World-mr & 12.0 & 138.7 & 42.9$\pm$19.5 & \bf{73.4$\pm$23.2}  & 57.1$\pm$10.7  & 17.4$\pm$8.1 &  \bf{75.2$\pm$23.1} & \bf{75.8$\pm$10.9}\\
 World-m & 58.1 & 206.8 & 74.6$\pm$11.5 & 93.8$\pm$31.8  & \bf{123.5$\pm$4.5}  & 73.4$\pm$9.3 &  119.3$\pm$18.2 & 119.7$\pm$16.7\\
 World-r & 0.0 & 63.1 & 5.9$\pm$5.2 & 68.0$\pm$20.8  & 2.0$\pm$3.0  & 2.8$\pm$5.5 &  98.3$\pm$49.7 & \bf{101.0$\pm$16.8}\\
 \hline
\end{tabular}
\label{table:results_mpe}
\end{table*}

\section{Experiments}

\subsection{Experimental Setup}\label{sec:experiment}

\textbf{Environments and Offline Dataset.} ~~ We consider various multi-agent environments categorized by their types, the number of agents, and levels of partial observability. Additionally, we use offline datasets generated by different MARL algorithms, covering various levels.

$(1)$ Three multi-agent particle environments (provided by \cite{pan2022plan} and generated using MA-TD3~\cite{ackermann2019reducing}): (a) Cooperative Navigation (CN), where three agents cover three landmarks without colliding, requiring coordination and teamwork; (b) Predator-Prey (PP), where three slower predators cooperate to capture a faster, pre-trained prey; and (c) World, where four predators attempt to catch two prey that aim to eat food while strategically using forest hiding spots for protection. For each task, evaluation is performed using four datasets of varying quality: expert, medium, medium-replay, and random. 

$(2)$ Three fully observable multi-agent MuJoCo environments (provided in \cite{wang2024offline} and generated using HAPPO~\cite{kuba2021trust}): We consider three tasks including 3-Agent Hopper (HC), 2-Agent Ant (Ant), and 6-Agent HalfCheetah (HC). For each task, evaluation is performed using four datasets of varying quality: expert, medium, medium-replay, and medium-expert.

$(3)$ Two partially observable multi-agent MuJoCo environments: 6-Agent HalfCheetah and 5-Agent Swimmer, each with differing degrees of partial observability and performance levels. We generated the offline datasets using ADER~\cite{kim2023adaptive} with varying levels of performance: expert, medium-1, medium-2, and their combinations. In both environments, each agent can observe its $K$ nearest neighbors. Here, $K$ determines the degree of partial observability. We consider $K = 0, 1$ for HC, which is denoted as HC-k$K$, and $K = 0$ for Swimmer (SW).

\textbf{Baselines.} ~~  We compare the proposed method against various baselines for each environment. For the multi-agent particle environments, we use MA-TD3+BC~\cite{pan2022plan}, OMAR~\cite{pan2022plan}, MADIFF~\cite{zhu2023madiff}, and CFCQL~\cite{shao2024counterfactual} as baselines. For multi-agent Mujoco environments, we use  OMAR~\cite{pan2022plan}, CFCQL~\cite{shao2024counterfactual}, and OMIGA~\cite{wang2024offline}. In addition, we include TD3+BC, FACMAC+BC, and TD3+B3C to analyze the effect of value factorization and \texttt{CC}.

\textbf{Implementation and Hyperparameters.} ~~To ensure a fair comparison, we implement the proposed method using the released CFCQL~\cite{shao2024counterfactual} code for the multi-agent particle environments and the released OMIGA~\cite{wang2024offline} code for the multi-agent Mujoco environments. The RL and BC coefficients in Eq. \ref{eq:facmac+b3c_policyupdate} are critical hyperparameters in our approach. As mentioned in Sec. \ref{sec:b3c_vf}, we initially set $\beta=1$ and tune $\alpha$ for each task, as done in \cite{fujimoto2021minimalist}. We observe that $\beta=1$ is sufficient for multi-agent Mujoco, but tuning $\beta$ improves performance in the multi-agent particle environments. For another hyperparameter, the scalar clipping value $M$ in Eq. \ref{eq:loss_cc}, we use $M=1$ for simplicity, except for the Ant task with the medium-replay dataset, where a lower $M$ significantly improves performance. An ablation study on this value is provided in Sec. \ref{sec:clipping_value}. The detailed hyperparameters used are listed in Appendix \ref{appendix:hyper}. 

\vspace{-2ex}

\begin{figure}[t]
\centering
\includegraphics[width=0.4\textwidth]{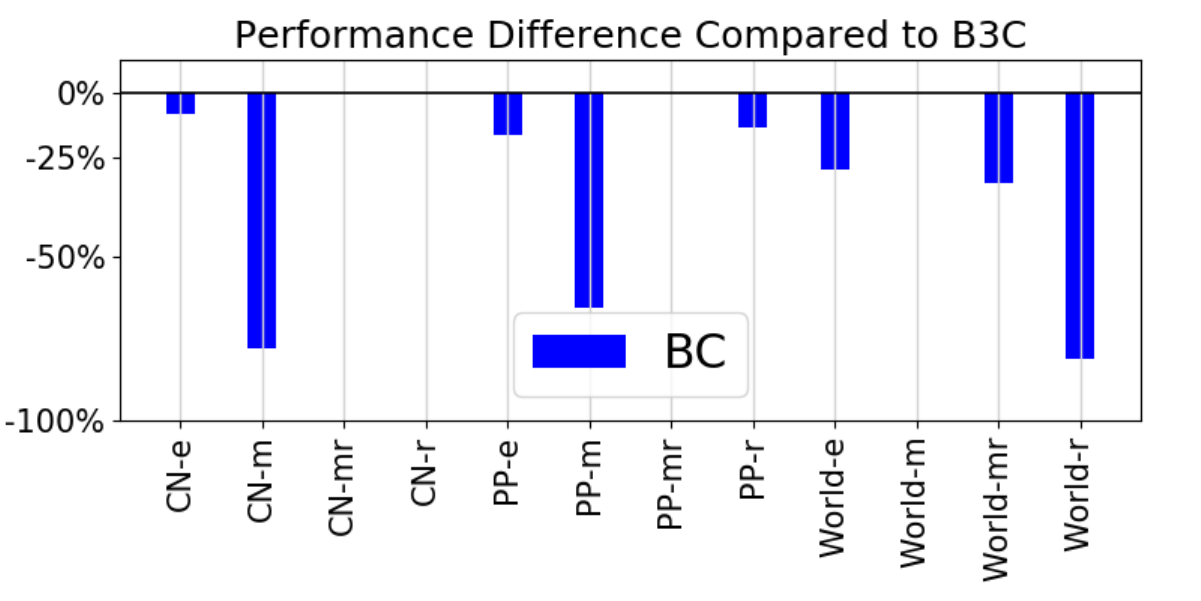}
\caption{Worst-seed performance gap between MA-TD3+BC and MA-TD3+B3C. Negative values indicate that MA-TD3+BC performs worse (i.e., a gap of $-5\%$ means its worst-seed performance is $5\%$ lower than that of MA-TD3+B3C).}
\label{fig:b3c_bc_min_difference_mpe}
\vspace{-4ex}
\end{figure}

\subsection{Numerical Results}\label{sec:numerical_results}

\newcolumntype{P}[1]{>{\centering\arraybackslash}p{#1}}

\begin{table*}[t!]
\footnotesize
\caption{Results of FACMAC+B3C and baselines on six Mujoco tasks across multiple datasets. Both maximum and average returns are reported. Dataset types are abbreviated as: expert (e), medium1 (m1), and medium2 (m2). Combined datasets are denoted as e-m1 and m1-m2. Bold indicates the best or comparable performance.}
\vspace{-3ex}
\vskip 0.1in
\centering
\renewcommand{\arraystretch}{1.2}
\begin{tabular}{|p{1.7cm}P{1cm}P{1cm}||P{1.cm}P{1.cm}P{1.cm}P{1.cm}P{1.7cm}|P{1.7cm}P{1.7cm}|}
 \hline
 \textbf{Task-Dataset}& Avg R & Max R &  \textbf{OMAR} &  \textbf{CFCQL}  & \textbf{OMIGA} & \textbf{TD3+BC} & \textbf{FACMAC+BC} & \textbf{TD3+B3C} & \textbf{FACMAC+B3C} \\
 \hline
 \hline
 \multicolumn{10}{|c|}{Fully observable multi-agent MuJoCo provided in OMIGA~\cite{wang2024offline} and generated by HAPPO~\cite{kuba2021trust}} \\
 \hline
 Hop-e & 2452.0 & 3762.7 & 2.4  & 718.5 & 859.6 & 3598 & 3621 & 1549 &\bf{3619.7$\pm$1.6}\\
 Hop-m & 723.6 & 2776.5 &  21.3  & 674.2 & 1189.3 & 1487 & 2880 & 1549 & \bf{3242.7$\pm$129.1}\\
 Hop-mr & 746.4 & 2801.2 & 3.3  & \bf{1380.2} & 774.2 & 330 & 143 & 263 & 736.8$\pm$469.4 \\
 Hop-me & 1190.6 & 3762.7 & 1.4  & 383.0 & 709.0  & 3212 & 3502 & 3356 & \bf{3328.0$\pm$369.8}\\
 \hline
 Ant-e & 2055.1 & 2124.2 & 313.5  & 1756.1 & 2055.5 & 2201& 2153& 2200& \bf{2162.8$\pm$46.0}\\
 Ant-m & 1418.7 & 1473.9  & -1710.0  & 1159.6 & 1418.4 & 1048 & 1042 & 1061 & \bf{1516.5$\pm$14.8}\\
 Ant-mr & 1029.5 & 1517.1  & -2014.2  & 1052.9 & 1105.1  & 959 & 968 & 1226 & \bf{1259.8$\pm$302.4}\\
 Ant-me & 1736.9 & 2124.2  & -2992.8  & 613.2 & 1720.3 & 2060 & 931 & 2174 & \bf{2077.6$\pm$194.2}\\
 \hline
 HC-e & 2785.1 & 3866.1  & -206.7  & 4999.2 &  3383.6 & 4777.2 & 1043.2 & 4724.3 & \bf{5403.5$\pm$169.7}\\
 HC-m & 1415.7 & 2113.5 & -265.7  & 4345.0 & 3608.1 & 2984.1 & 4667.7 & 2995.2 & \bf{4756.6$\pm$56.3}\\
 HC-mr & 655.8 & 2132.6 & -235.4  & 3655.3 & 2504.7  & 3652.7 & 4538.6 & 3635.9 & \bf{4602.6$\pm$150.2}\\
 HC-me & 2105.4 & 3866.1 &  -253.8  & 5030.9 & 2948.5  & 4826.3 & 5395.2 & 4860.6 & \bf{5413.7$\pm$99.4}\\
 \hline
 \hline
 \multicolumn{10}{|c|}{Partial observable multi-agent MuJoCo generated by ADER~\cite{kim2023adaptive}} \\
 \hline
 HC-k0-e & 1394.3 & 1520.8 &  197.2  & 750.7 &  1390.1 & -145 & 1381 & 1316 & \textbf{1396.8$\pm$4.5}\\
 HC-k0-m1 & 1103.3 & 1332.9 & 189.6  & 443.7 &  1106.3 & 1079 & 1158 & 1078 & \textbf{1141.6$\pm$18.9}\\
 HC-k0-m2 & 840.8 & 1157.4 & 839.1  & 766.5 &  847.6 & 967 & 1197 & 963 & \textbf{1195.0$\pm$51.9}\\
 HC-k0-e-m1 & 1252.3 & 1520.8 &  136.4 & 542.6 &  1199.5 & 590 & 784 & 1209 & \textbf{1307.1$\pm$27.3}\\
 HC-k0-e-m2 & 1121.7 & 1520.8 &  299.2  & 398.4 &  1097.9 & 98 & 1211 & 1080 & \textbf{1230.0$\pm$40.5}\\
 HC-k0-m1-m2 & 976.2 & 1332.9 &  709.1  & 1186.1 &  1027.3 & 1007 & 1107 & 995 & \textbf{1210.1$\pm$22.8}\\
 \hline
 HC-k1-e & 3766.0 & 3863.3 &  3232.6  & 3390.2 & 3089 & 3731 & 3748 & \textbf{3760.8} & \textbf{3760.5$\pm$24.2}\\
 HC-k1-m1 & 1976.2 & 2350.0 &  2312.4  & 2356.0 &  2017.3 & 2182 & 2413 & 2200 & \textbf{2508.1$\pm$55.7}\\
 HC-k1-m2 & 1223.9 & 1758.4 &  1108.9  & 1440.6 &  1196.5 & 1244 & 1284 & 1387 & \textbf{2187.8$\pm$66.7}\\
 HC-k1-e-m1 & 2873.2 & 3863.3 &  2296.4  & 1949.6 &  2112.3 & 1892 & 2611 & 2380 & \textbf{2682.0$\pm$175.4}\\
 HC-k1-e-m2 & 2486.0 & 3863.3 &  524.8  & 864.7 &  \textbf{1088.2} & 992 & 1106 & 1211 & \textbf{1102.8$\pm$349.8}\\
 HC-k1-m1-m2 & 1591.6 & 2237.2 &  1410.6  & 1862.2 &  1633.5 & 1926 & 1186 & 2021 & \textbf{2222.6$\pm$84.1}\\
 \hline
 Sw-e & 430.9 & 438.9 & 395.3  & 403.3 &  \textbf{430.7} & 371.0 & 424.1 & 428.2 &\textbf{430.3$\pm$2.6}\\
 Sw-m1 & 290.4 & 306.3 & 268.4  & 277.1 &  288.3 & 295.6 & 288.8 & \textbf{297.0} &289.5$\pm$1.1\\
 Sw-m2 & 142.7 & 158.5 & 97.9  & 130.0 &  142.6 & 140.1 & 76.6 & 147.1 &\textbf{155.3$\pm$1.7}\\
 Sw-e-m1 & 360.6 & 438.9 & 216.5  & 292.1 &  261.4 & 68.6 & 99.9 & 288.4 &\textbf{297.1$\pm$53.8}\\
 Sw-e-m2 & 286.8 & 438.9 & 113.8  & 138.6 &  148.2 & 160.8 & 213 & \textbf{236.1} & 211.9$\pm$8.4\\
 Sw-m1-m2 & 216.3 & 306.3 &  226.4  & 222.7 &  205.8 & 185.4 & 200.7 & \textbf{228.4}  &189.9$\pm$7.7\\
 \hline
\end{tabular}
\label{table:results_mujoco}
\end{table*}

\textbf{Multi-agent particle environments.} ~~ The results are shown in Table \ref{table:results_mpe}. As discussed in Sec. \ref{sec:overreg}, MA-TD3+BC from prior works suffers from over-regularization, which is evident from its performance on medium-replay and random datasets, where it significantly underperforms compared to the baselines. Thus, we first further tune the coefficients of RL and BC, which is referred to as MA-TD3+BC (ours) in Table \ref{table:results_mpe}. This simple tuned version substantially improves performance and even outperforms the baselines in some cases. However, this simple hyperparameter tuning is insufficient: \textit{BC regularization alone is often unstable}. We claim that the proposed \texttt{CC} can address this instability. Fig. \ref{fig:b3c_bc_min_difference_mpe} illustrates the performance difference between the worst-performing seed among 5 seeds of MA-TD3+BC and that of MA-TD3+B3C, where $-5\%$ indicates that the worst-performing seed of MA-TD3+BC performs $5\%$ worse than MA-TD3+B3C. As shown in Fig. \ref{fig:b3c_bc_min_difference_mpe}, using BC alone results in a significant performance drop in the worst-performing seed compared to B3C due to its instability, demonstrating that B3C, which incorporates critic clipping, stabilizes learning. Additionally, in terms of average performance, TD3+B3C outperforms the baselines in most cases. In particular, on the medium-replay and random datasets, which are generated by multiple behavior policies, TD3+B3C achieves state-of-the-art performance with large margins.

\textbf{Multi-agent Mujoco.} ~~In this environment, we evaluate FACMAC+B3C, which combines non-monotonic value factorization with B3C. For fully observable cases, we use the reported results of OMAR and OMIGA~\cite{wang2024offline}, and generate results for TD3+BC, FACMAC+BC, and TD3+B3C for comparison. As shown in Table~\ref{table:results_mujoco}, FACMAC+B3C achieves the best overall performance across environments, often exceeding the dataset’s maximum return, particularly on HalfCheetah and several Hopper and Ant tasks. Moreover, unlike the baselines, FACMAC+B3C exhibits no performance collapse across any dataset, indicating stable and consistent learning.

The consistent improvement of FACMAC+B3C over FACMAC+BC, as well as TD3+B3C over TD3+BC, demonstrates the effectiveness of \texttt{Critic Clipping} in stabilizing value estimation and improving overall policy quality. This result highlights that the benefit of B3C is not limited to a specific architecture but generalizes across both factored and non-factored critics.

Furthermore, FACMAC+BC generally outperforms TD3+BC except for a few fully observable tasks such as Ant-me and HC-e, which aligns with findings from online MARL literature where value factorization improves coordination and representation efficiency. These results collectively confirm that integrating value factorization with B3C provides both high and stable performance in complex multi-agent continuous control settings.

\vspace{-1ex}

\begin{figure*}[t]
\hspace{5ex}\includegraphics[width=0.2\linewidth]{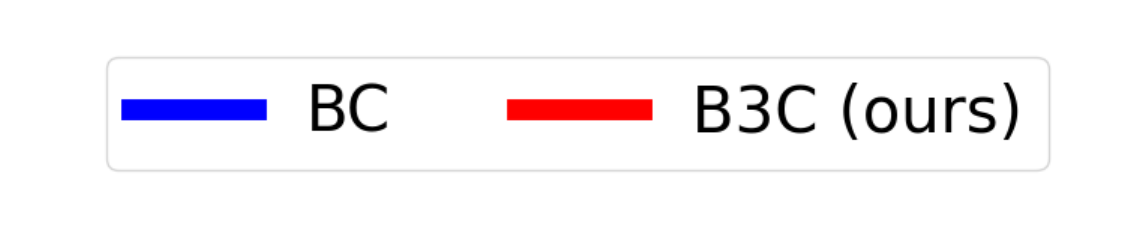} \\
\begin{tabular}{cccc}
\hspace{-2ex}\includegraphics[width=0.24\linewidth]{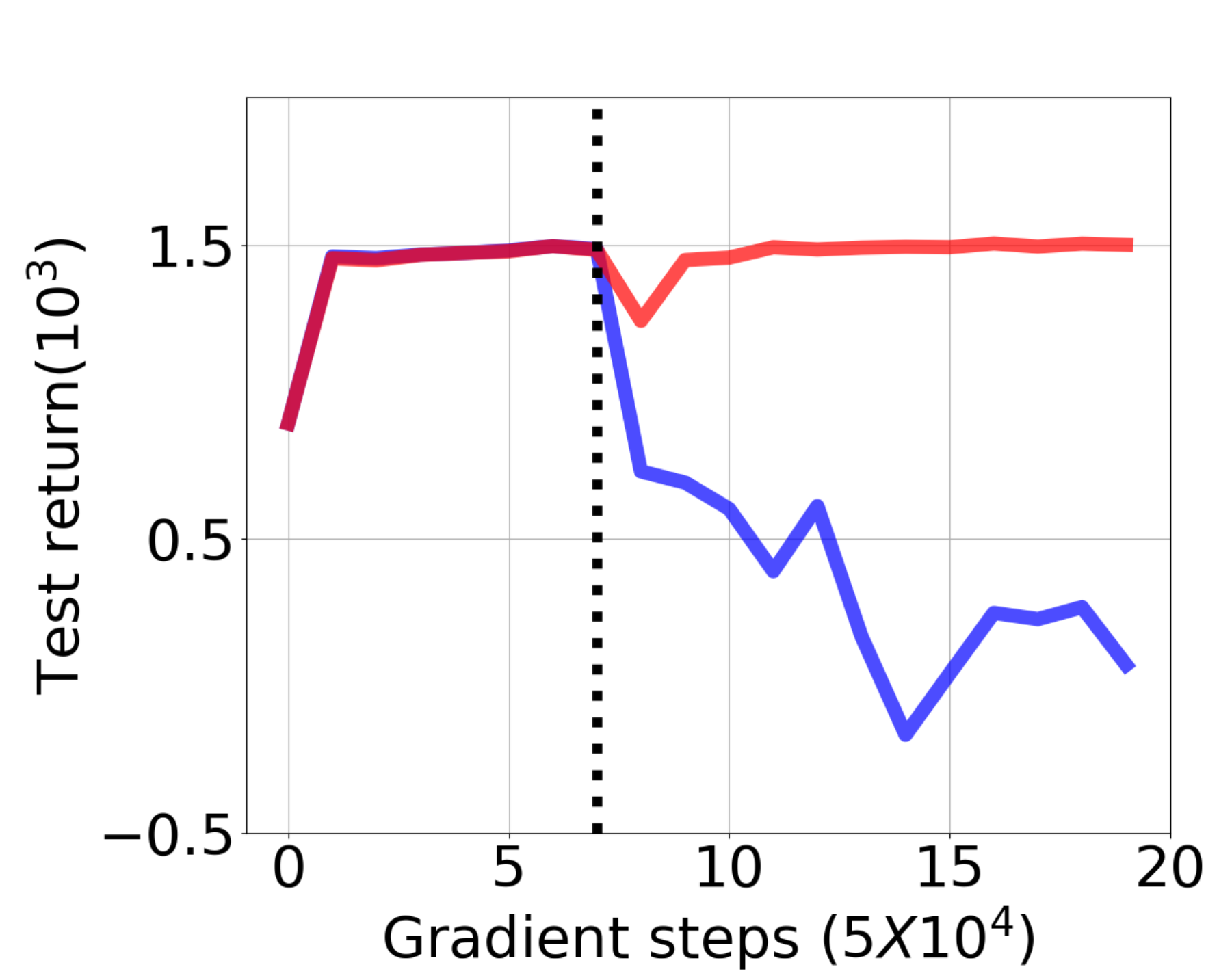} &
\hspace{-2ex}\includegraphics[width=0.24\linewidth]{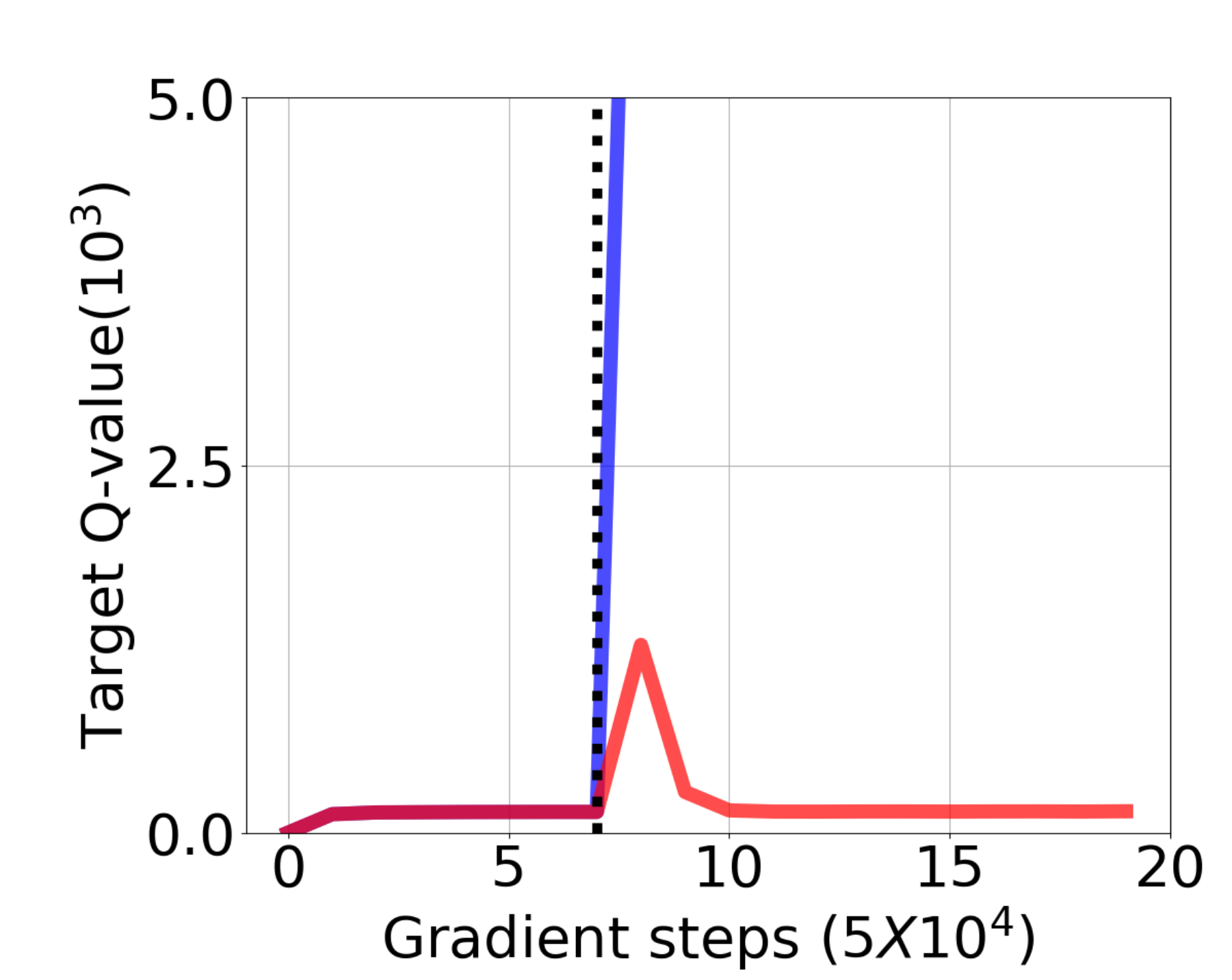} &
\hspace{-2ex}\includegraphics[width=0.24\linewidth]{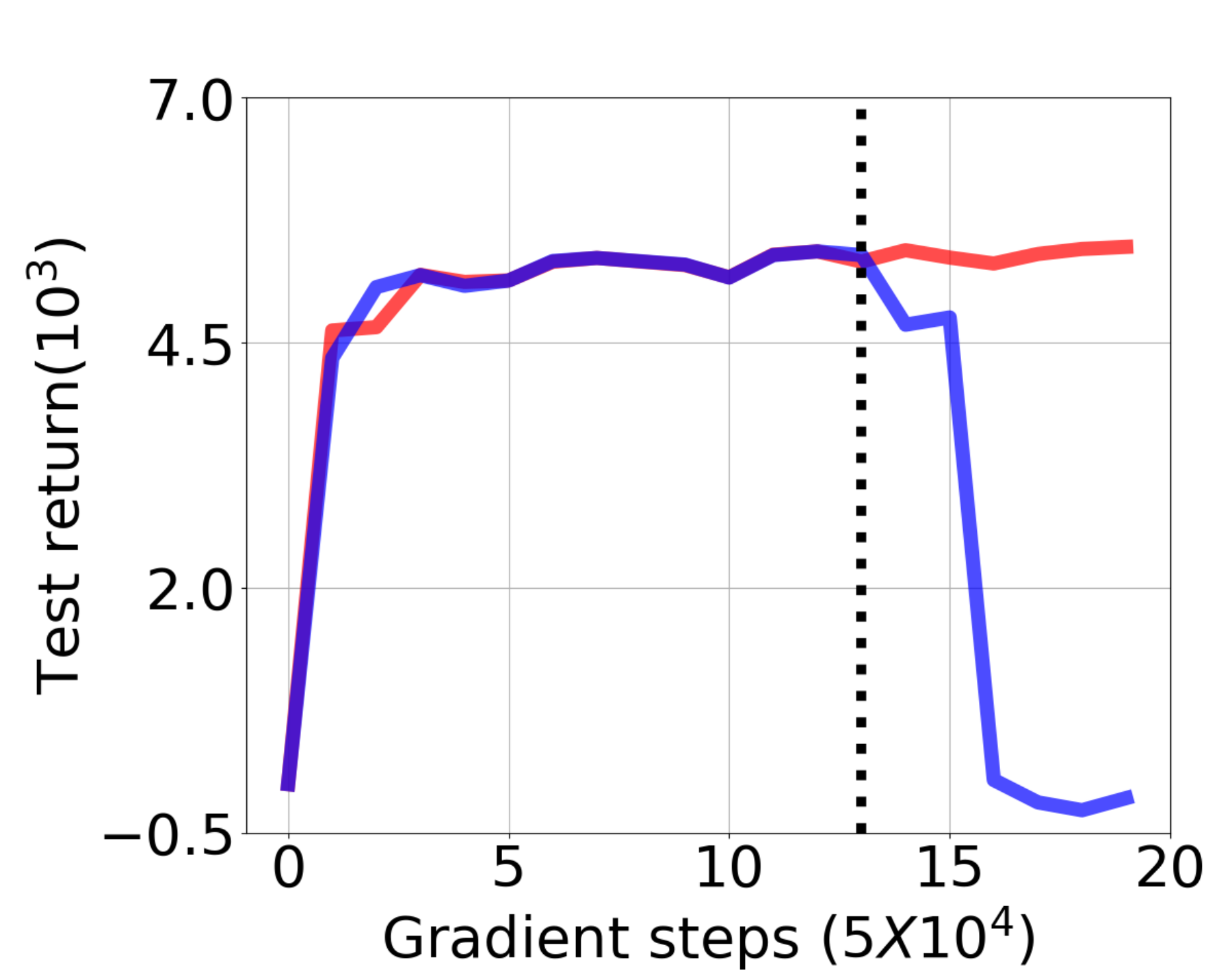} &
\hspace{-2ex}\includegraphics[width=0.24\linewidth]{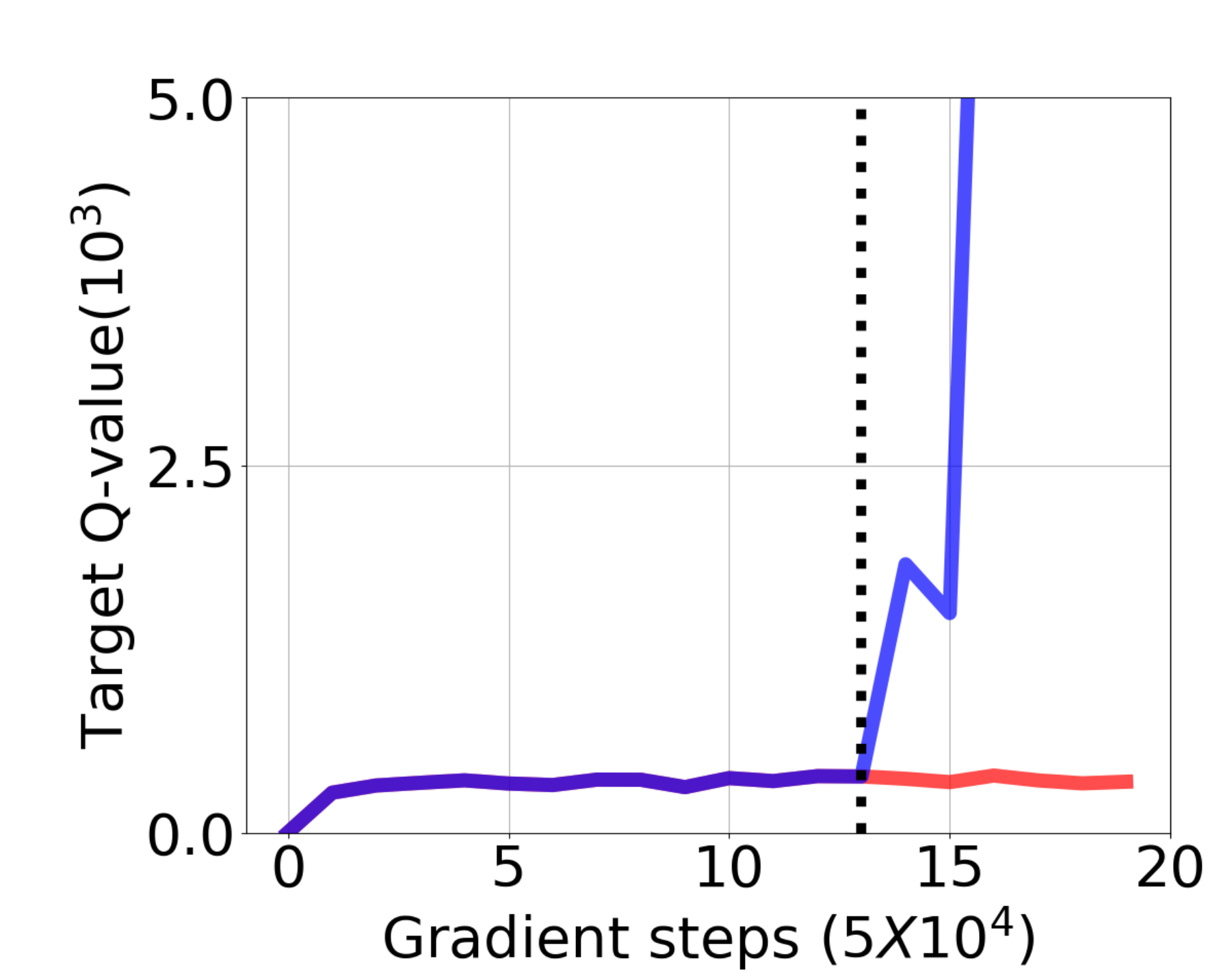} \\
(a) Ant2x4-return & (b) Ant2x4-target & (c) HC6x1-return & (d) HC6x1-target \\
\end{tabular}
\caption{Test return and target value during policy evaluation in the training of BC (blue) and B3C (red). The black dotted line indicates the moment when the target value starts to diverge.}
\label{fig:cc}
\end{figure*}

\begin{figure*}[t]
\hspace{4ex}\includegraphics[width=0.2\linewidth]{figures/b3c_bc_legend.pdf} \ \\
\begin{tabular}{cccccc}
\hspace{-3ex}\includegraphics[width=0.165\linewidth]{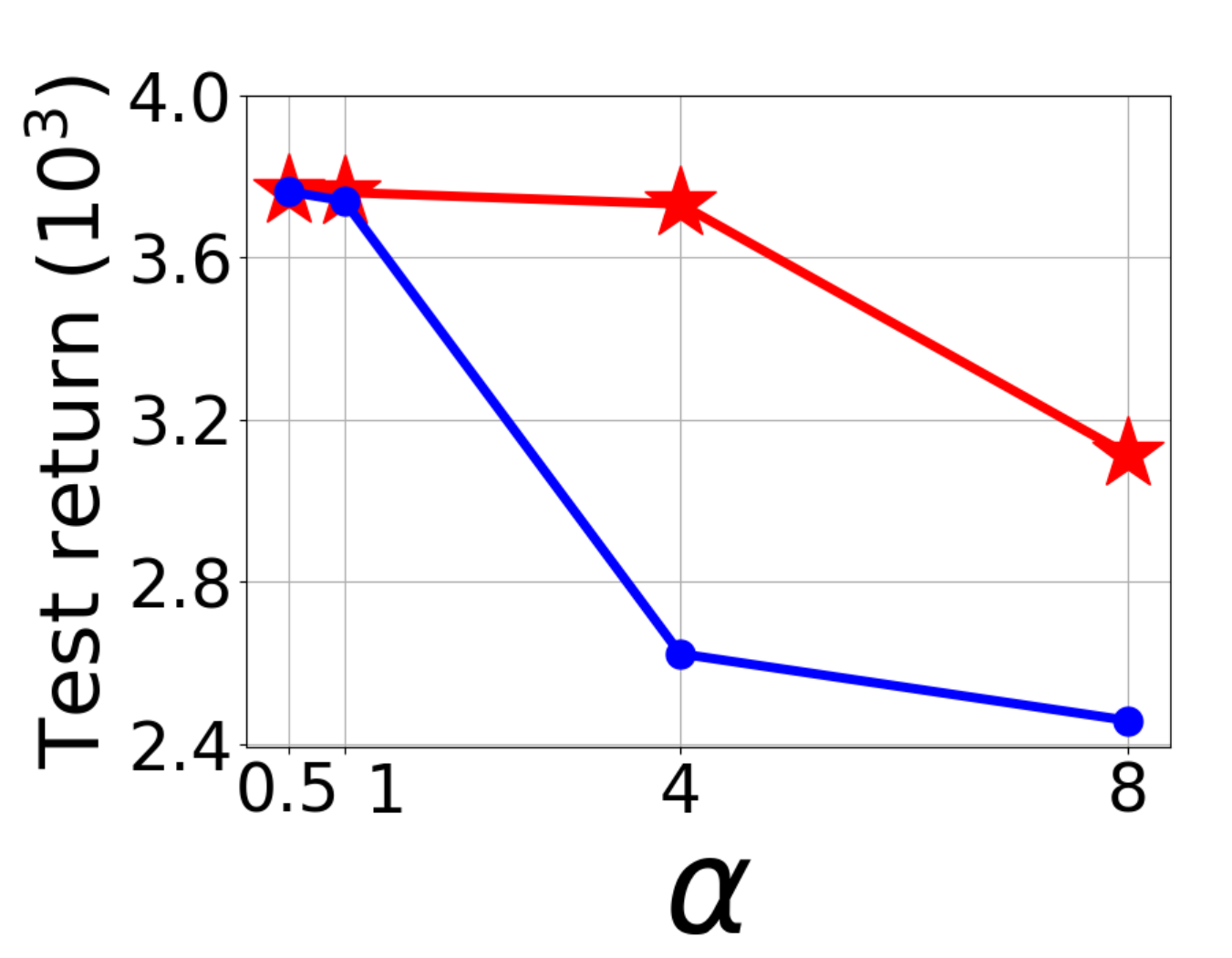} & 
\hspace{-3ex}\includegraphics[width=0.165\linewidth]{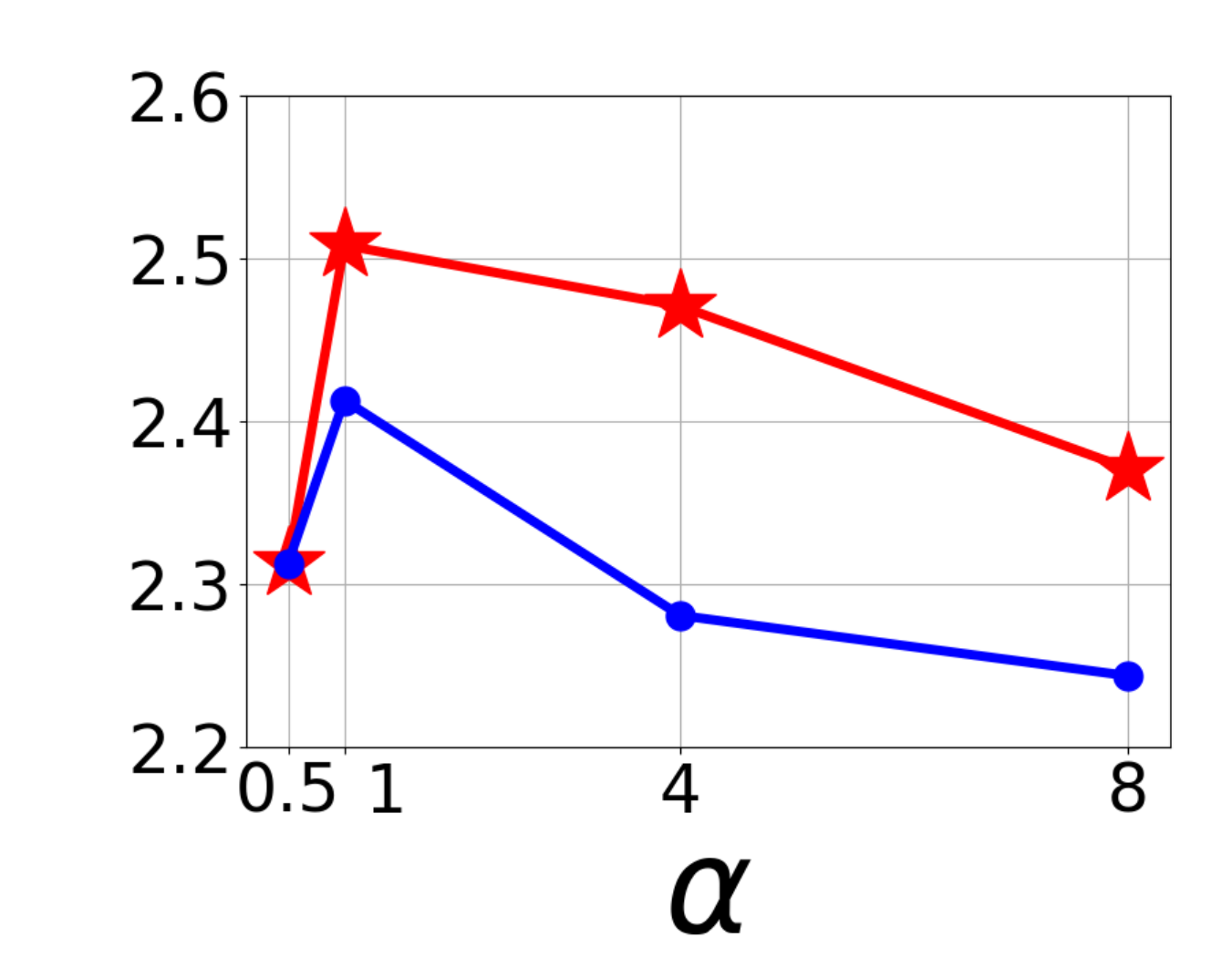} &\hspace{-3ex}
\includegraphics[width=0.165\linewidth]{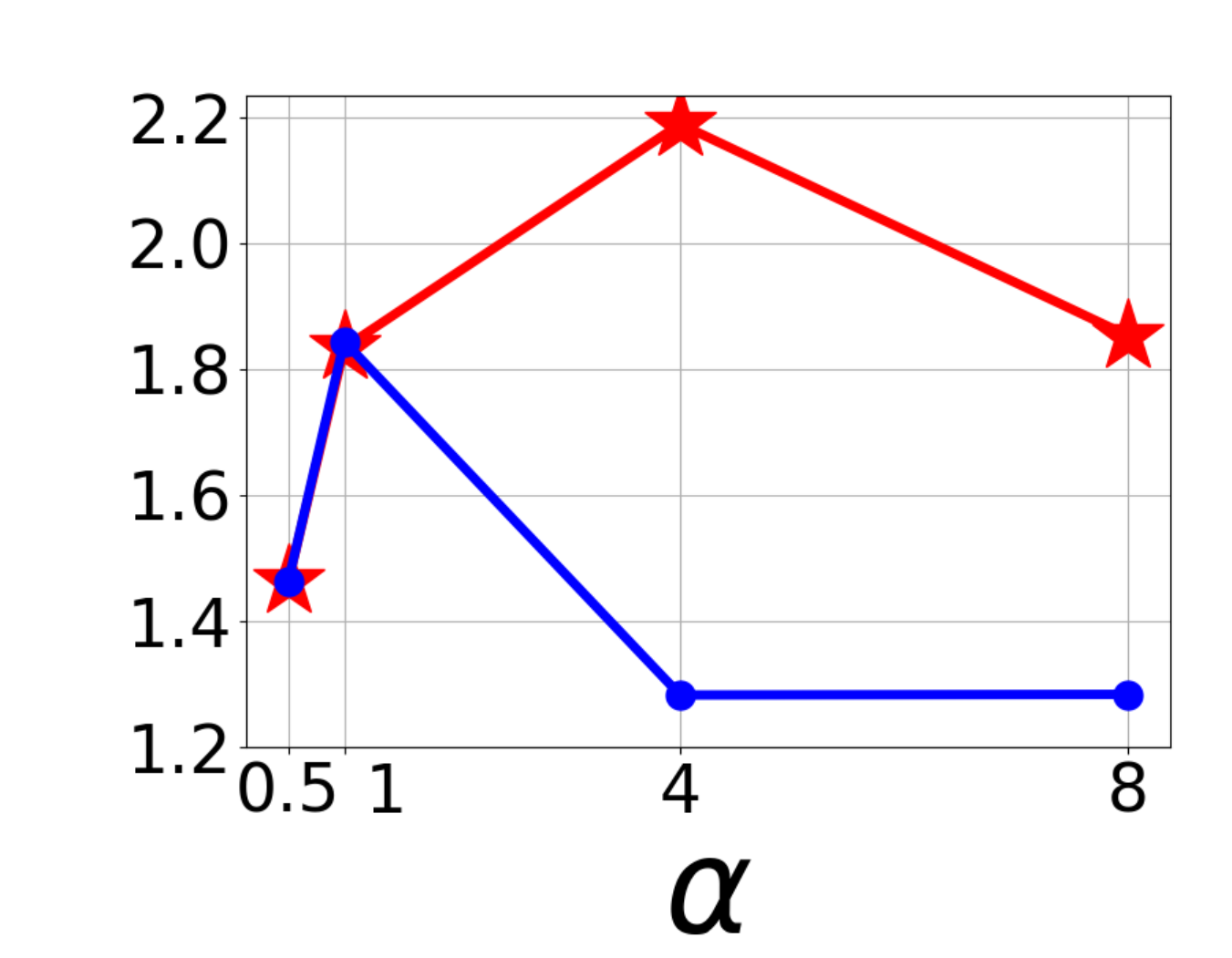} &\hspace{-4ex}
\includegraphics[width=0.165\linewidth]{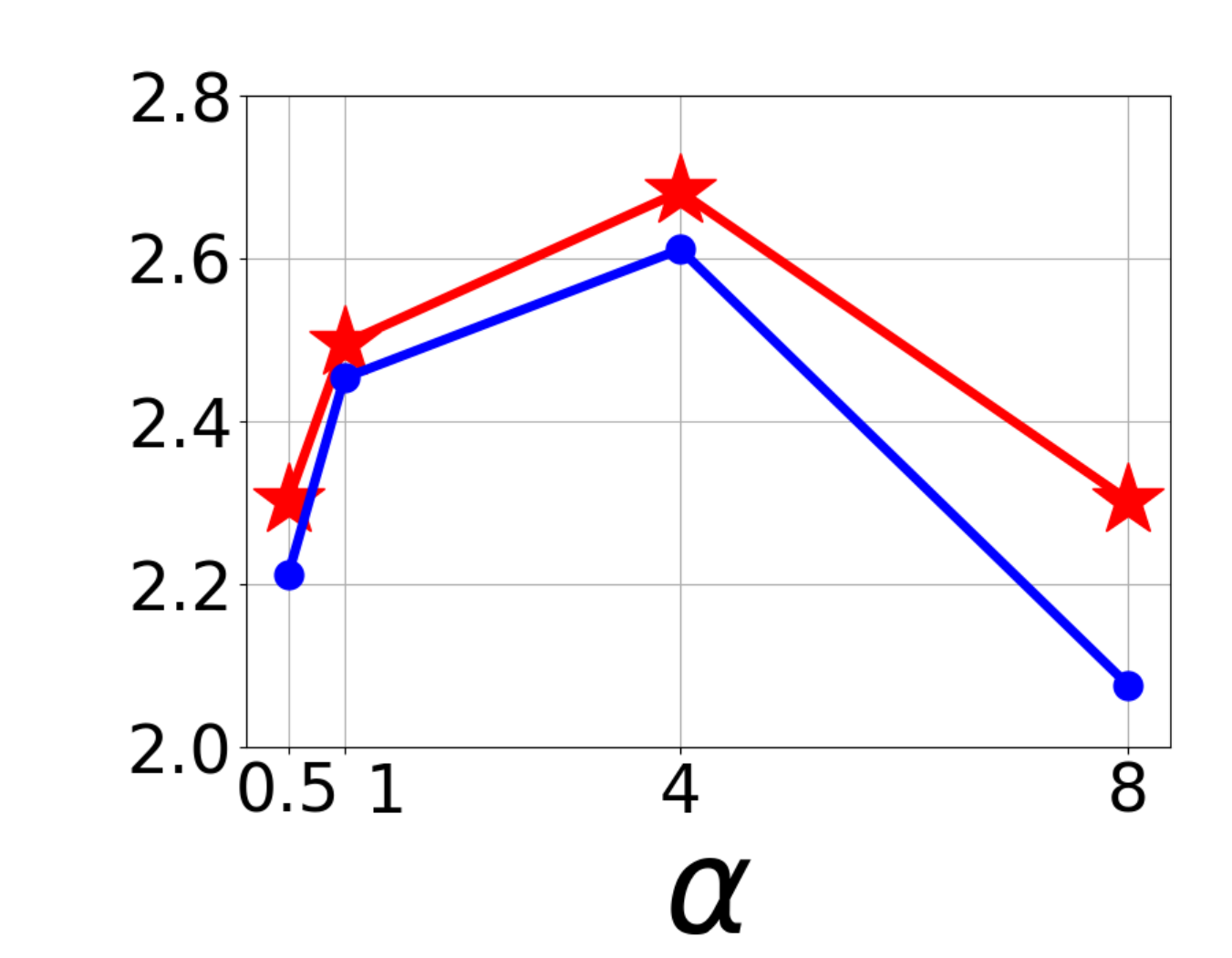} &\hspace{-4ex}
\includegraphics[width=0.165\linewidth]{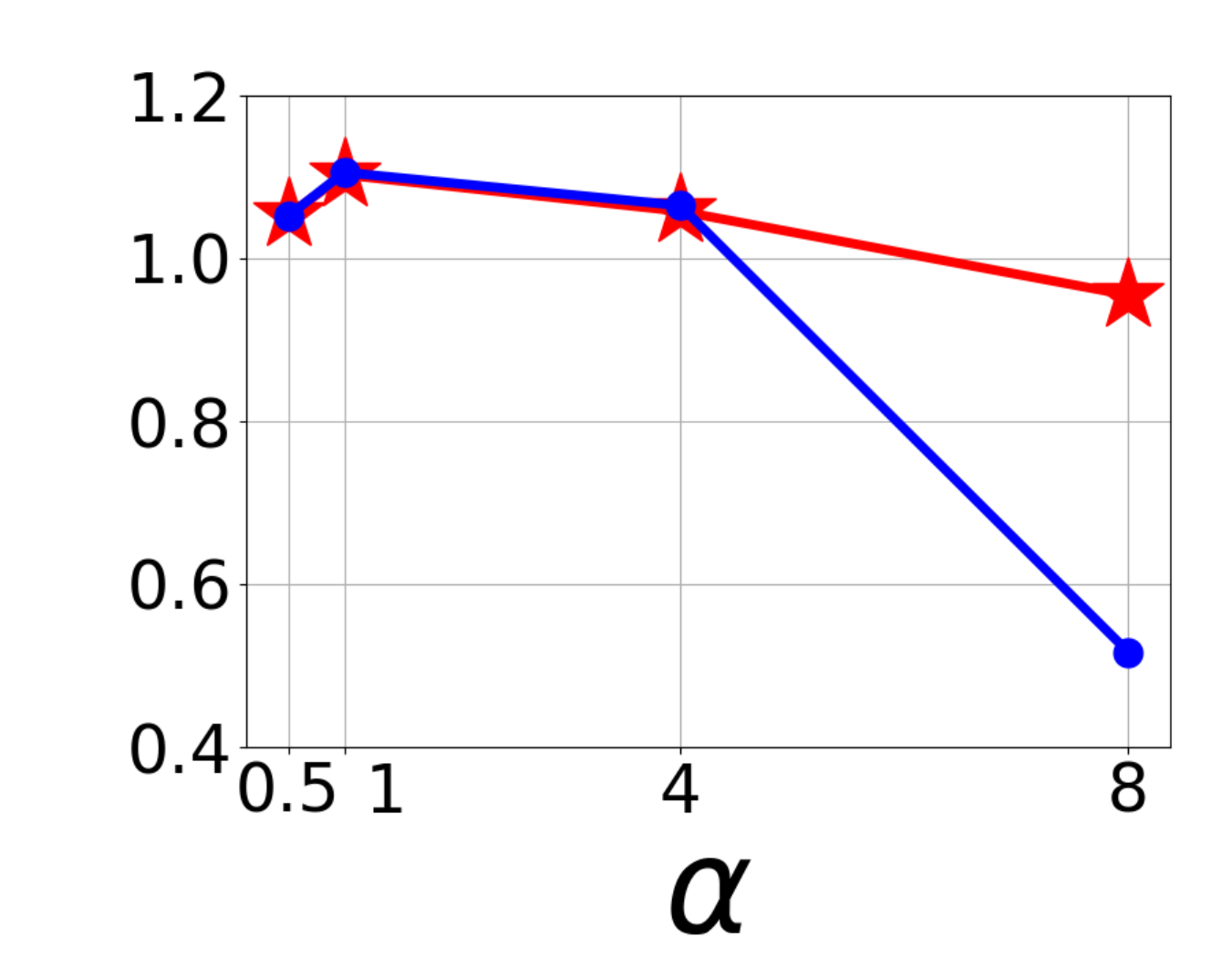} &\hspace{-4ex}
\includegraphics[width=0.165\linewidth]{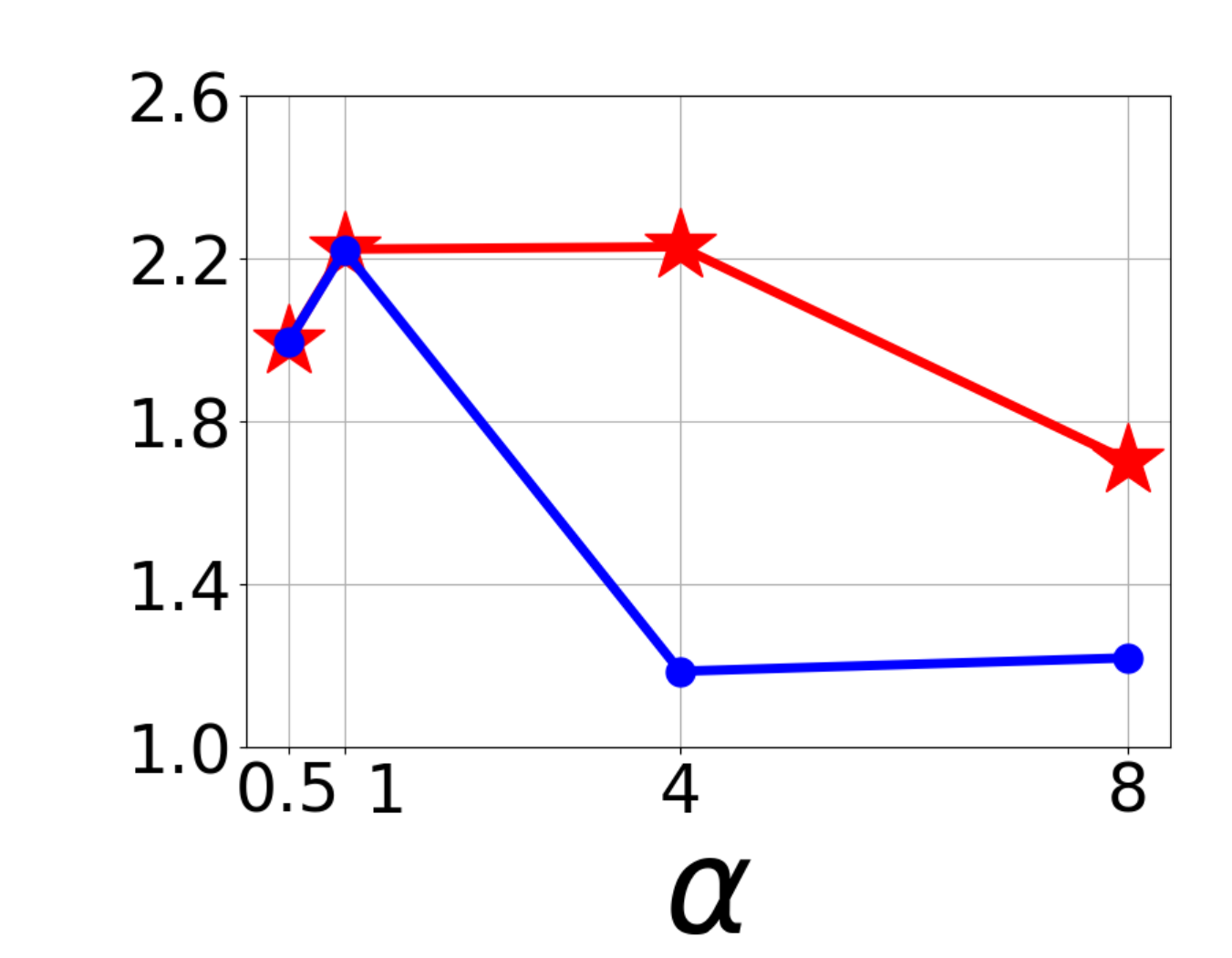} \\
\hspace{-3ex}(a) HC-k1-e & \hspace{-2ex}(b) HC-k1-m1 & \hspace{-2ex}(c) HC-k1-m2 & \hspace{-2ex}(d) HC-k1-e-m1 & \hspace{-2.5ex}(e) HC-k1-e-m2 & \hspace{-1ex}(f) HC-k1-m1-m2
\end{tabular}
\caption{Performance with respect to the RL coefficient $\alpha$ in the partially observable Halfcheetah environment. Larger $\alpha$ increases the weight on the RL objective relative to BC regularization.}
\label{fig:perfor_alpha}
\end{figure*}

\begin{figure}[t]
\centering
\includegraphics[width=0.35\textwidth]{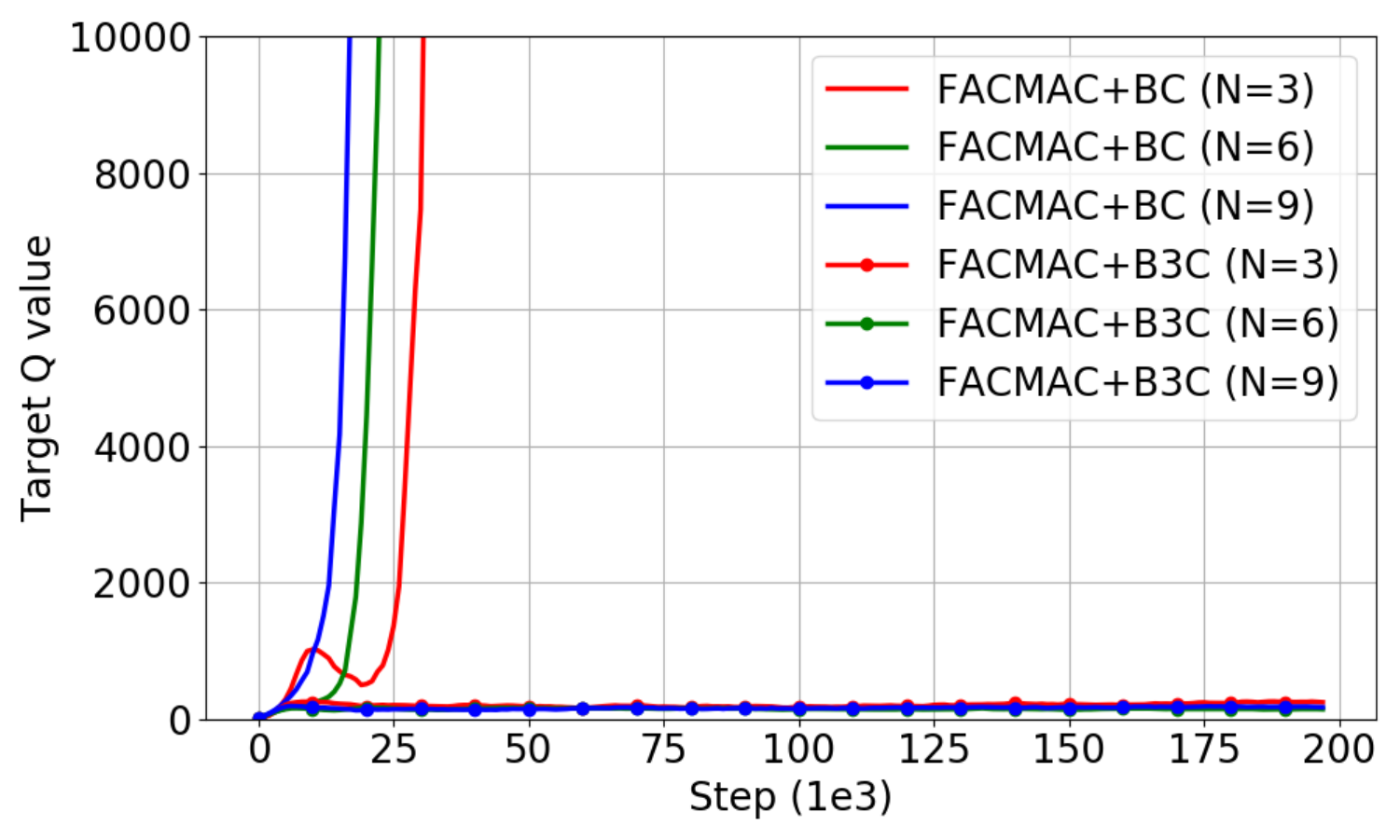}
\vspace{1ex}

\begin{minipage}{\linewidth}
\small
\centering
\begin{tabular}{lccc}
\toprule
Method & PP (N=3) & PP (N=6) & PP (N=9) \\
\midrule
FACMAC+BC & 71.1 $\pm$ 3.6 & 74.8 $\pm$ 20.1 & 85.6 $\pm$ 47.8 \\
FACMAC+B3C & 97.1 $\pm$ 6.0 & 90.8 $\pm$ 12.4 & 104.5 $\pm$ 16.8 \\
\bottomrule
\end{tabular}
\end{minipage}

\caption{Performance and target Q-values of FACMAC with BC and B3C for varying $N$.}
\label{fig:b3c_pp_target_369}
\end{figure}

\subsection{Analysis}

\subsubsection{Critic Clipping: B3C versus BC}\label{sec:b3c_versus_bc} 

\textbf{Critic clipping alleviates overestimation} by directly constraining the target value in policy evaluation. As mentioned earlier, BC regularization alone often makes learning unstable, and even the critic diverges. We have already shown instability problem in Sec. \ref{sec:numerical_results} through the worst performance of MA-TD3+BC in the multi-agent particle environments. Here, we provide further analysis of how critic clipping addresses this issue for better understanding in multi-agent Mujoco environments.

Fig. \ref{fig:cc} presents the test return and the target value, $y^{jt}$ from Eq. \ref{eq:loss_cc}, during the training of FACMAC+BC and FACMAC+B3C. The results indicate that FACMAC+BC, which relies solely on BC regularization, results in diverging value estimates. When this divergence occurs, marked by the dotted line in Fig. \ref{fig:cc}, the performance correspondingly degrades. However, B3C effectively mitigates this divergence---it occasionally tends to overestimate but either stabilizes quickly or avoids divergence entirely, ensuring more consistent performance. Note that this phenomenon does not always occur but is more likely when the RL objective is weighted more heavily than BC regularization; however, critic clipping can prevent it and enable stable learning. 

Additionally, we evaluate FACMAC+BC and FACMAC+B3C in a partially observable Predator-Prey environment~\cite{peng2021facmac} with 3, 6, and 9 agents. This experiment aims to examine how the degree of overestimation changes as the number of agents $N$ increases. For a fair comparison, we normalize each dataset return by its average to compare target Q-values across the 3-, 6-, and 9-agent settings. As shown in Fig.~\ref{fig:b3c_pp_target_369}, the target Q-values of FACMAC+BC diverge more rapidly as $N$ increases, indicating that overestimation becomes more severe with a larger number of agents due to the exponentially growing joint action space. In contrast, critic clipping in FACMAC+B3C effectively suppresses this divergence, keeping target values bounded even under increased coordination complexity. This stability corresponds to higher and more consistent performance, as shown in the table below, indicating that B3C maintains robustness as the number of agents increases and alleviates the instability induced by large-scale coordination in multi-agent offline RL.

\begin{figure*}[t!]
\hspace{6ex}\includegraphics[width=0.25\linewidth]{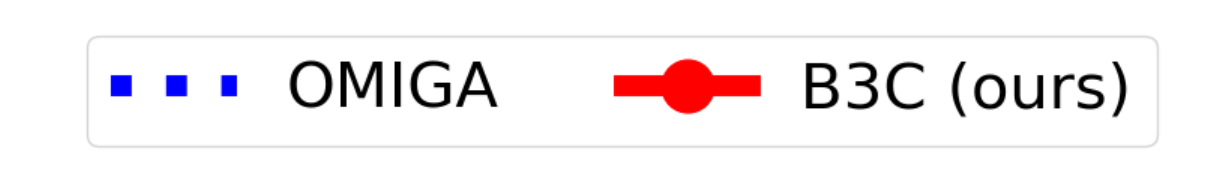} \\
\begin{tabular}{cccccc}
\hspace{-2ex}\includegraphics[width=0.165\linewidth]{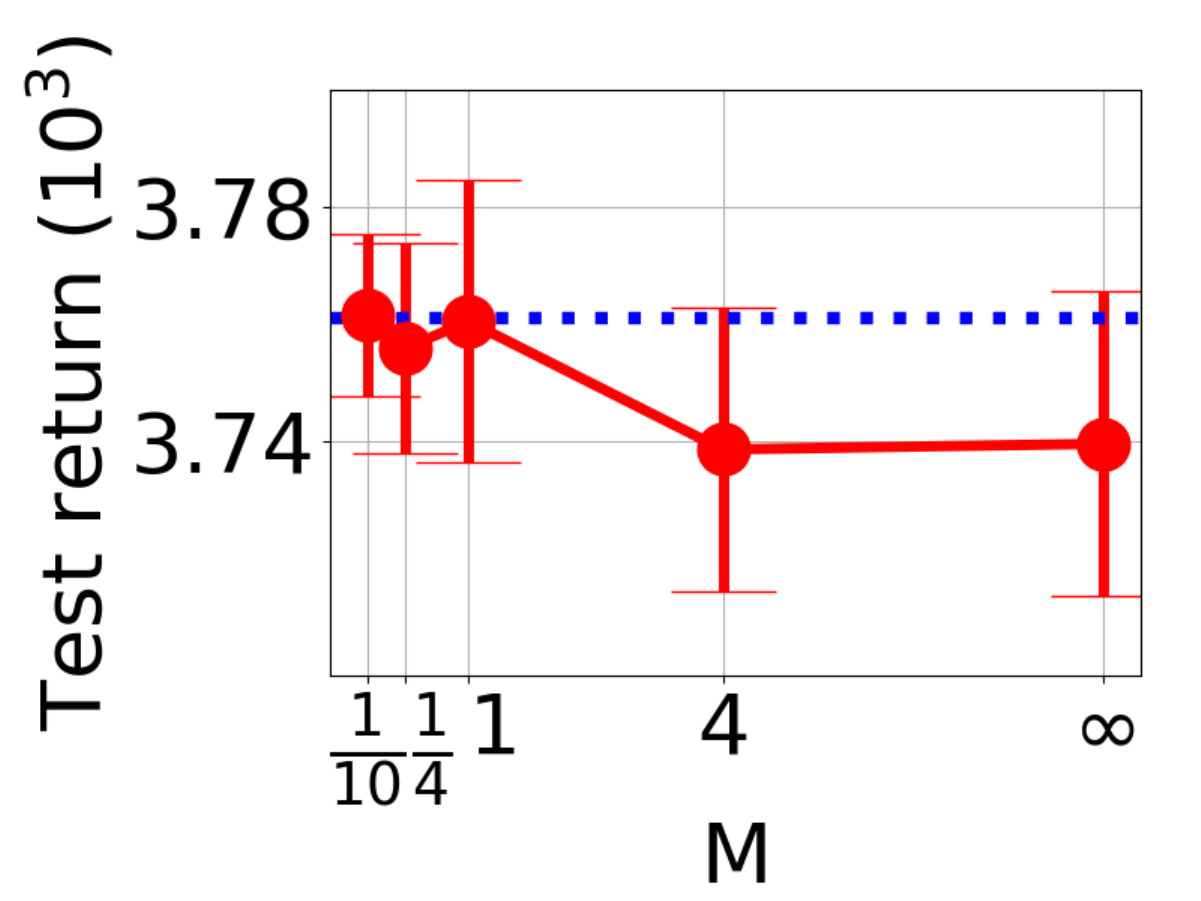} & \hspace{-3ex}
\includegraphics[width=0.165\linewidth]{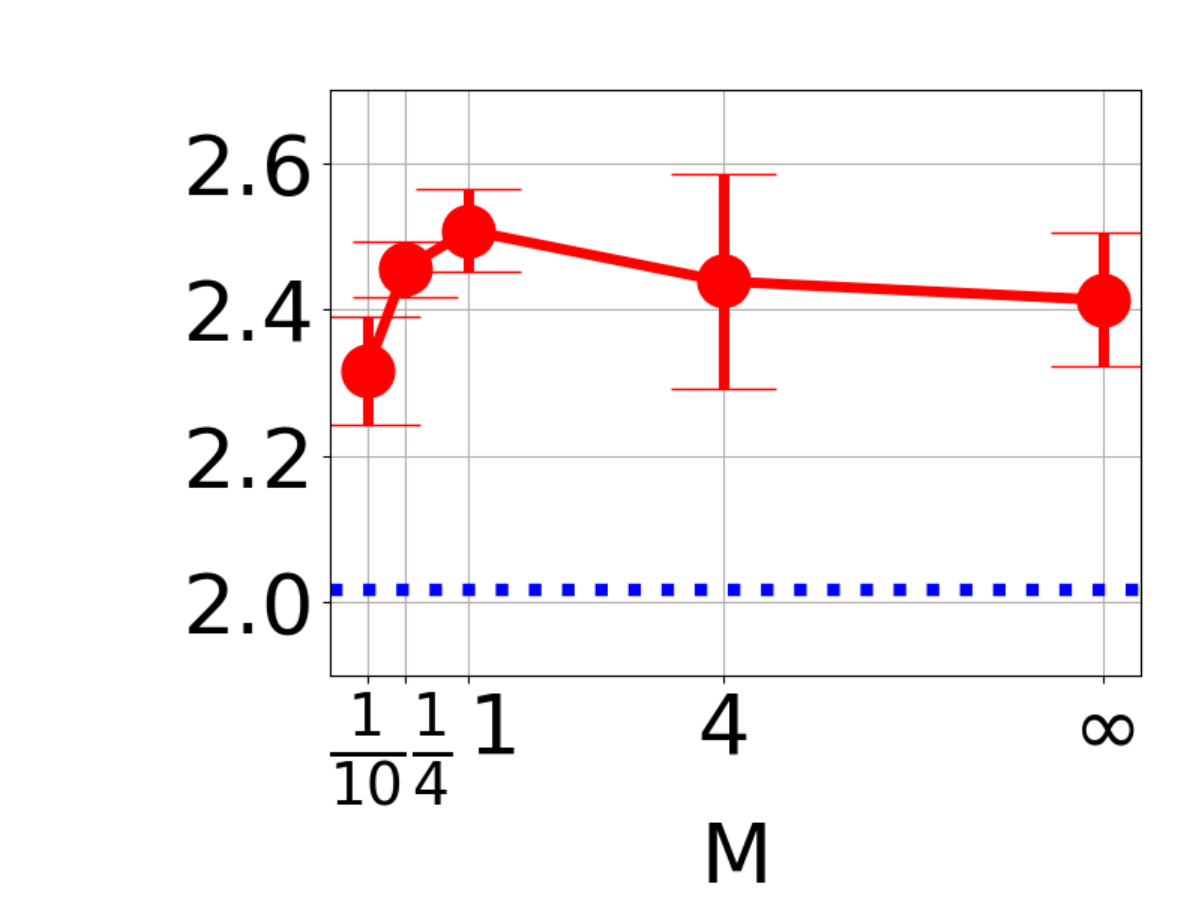} &\hspace{-3ex}
\includegraphics[width=0.165\linewidth]{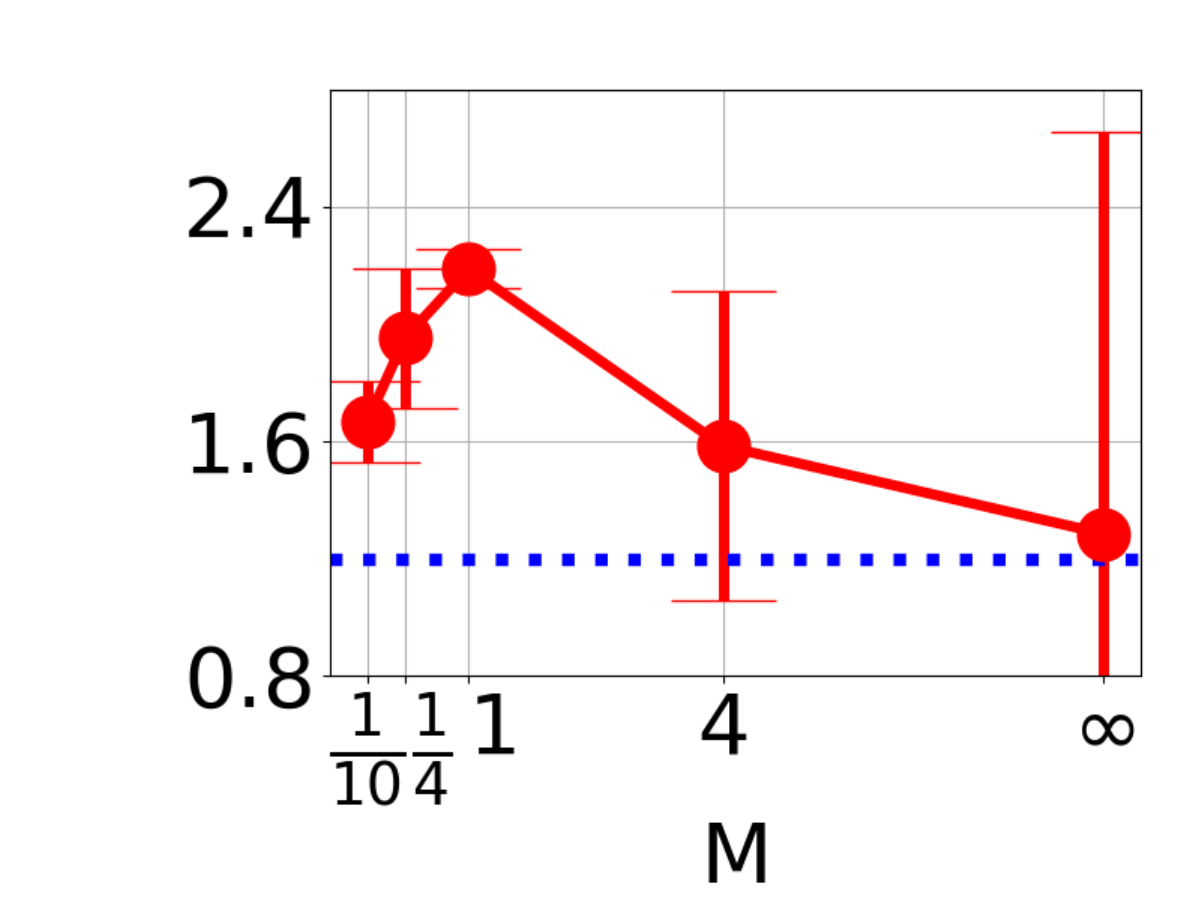} &\hspace{-5ex}
\includegraphics[width=0.165\linewidth]{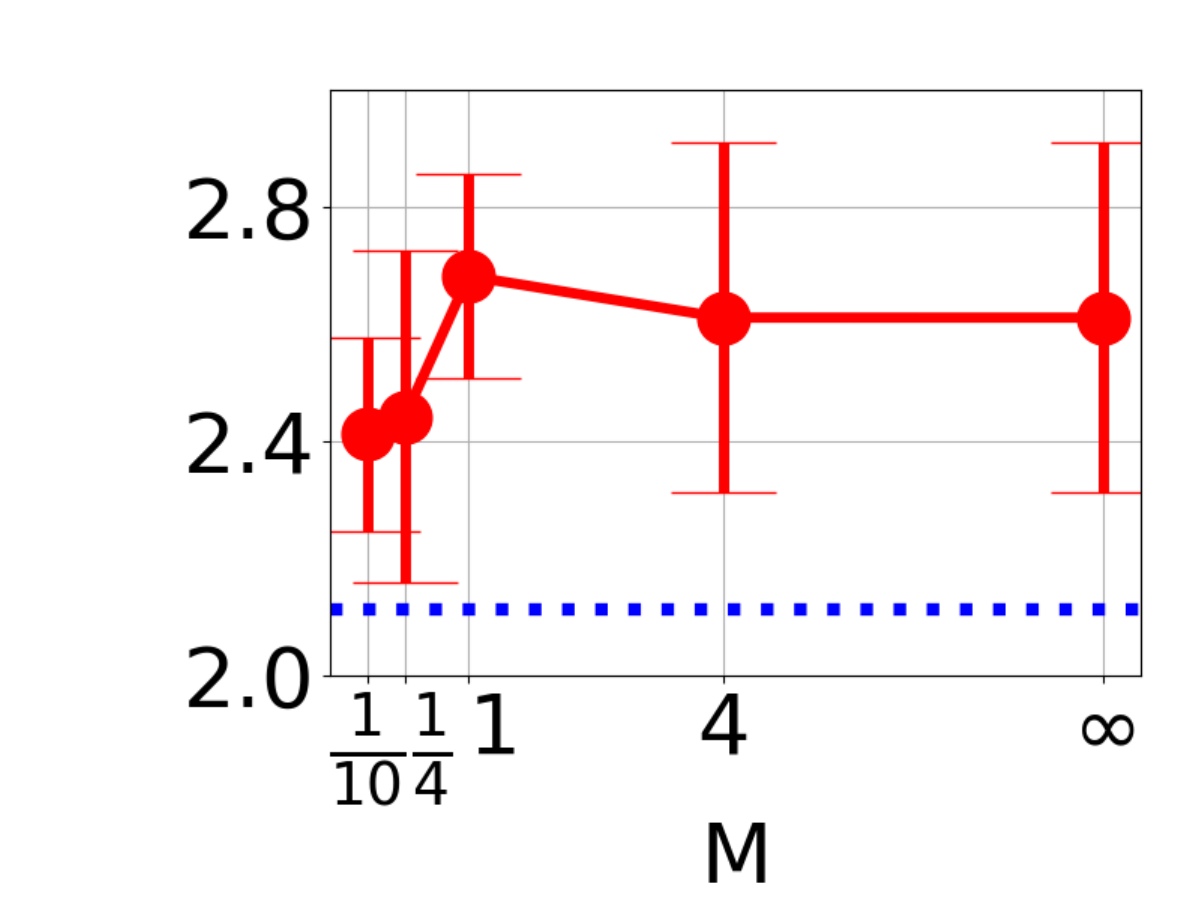} &\hspace{-6ex}
\includegraphics[width=0.165\linewidth]{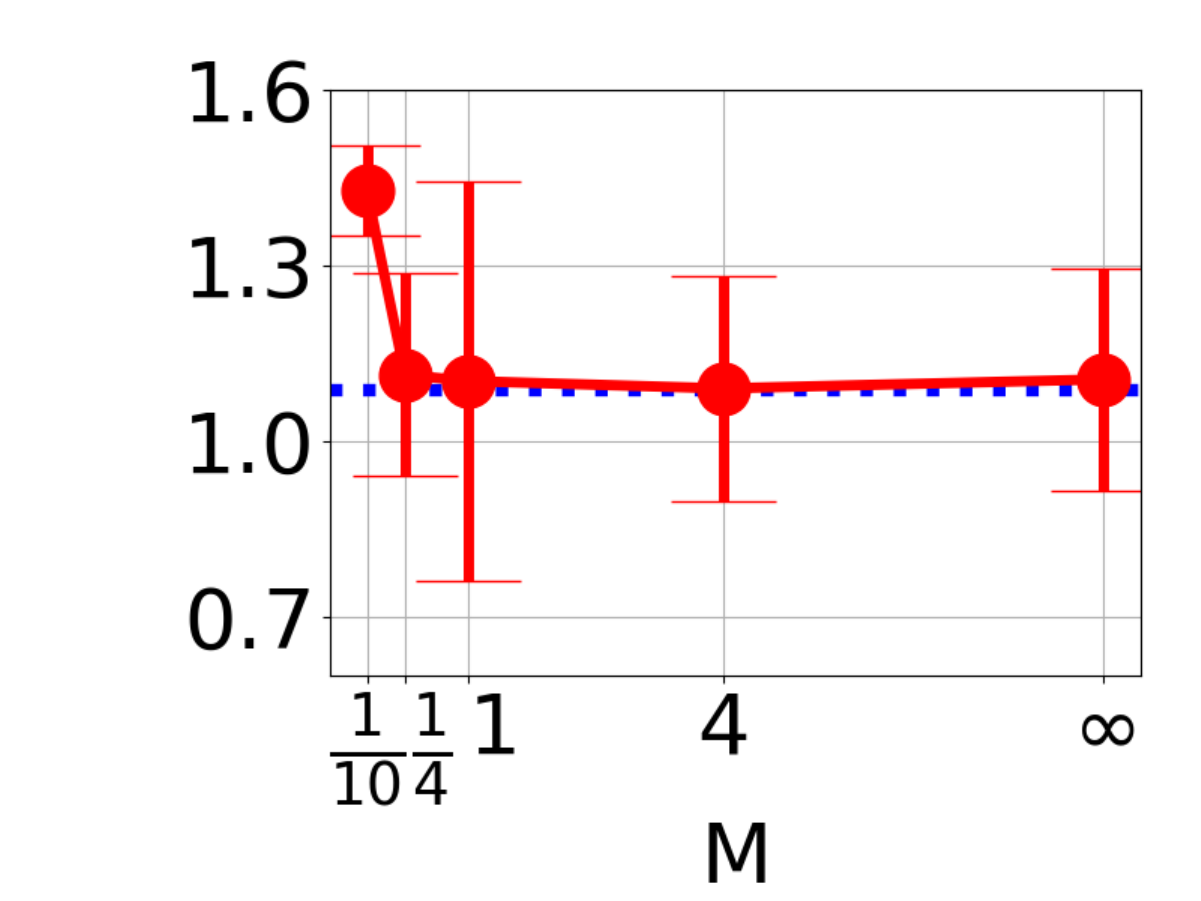} &\hspace{-8ex}
\includegraphics[width=0.165\linewidth]{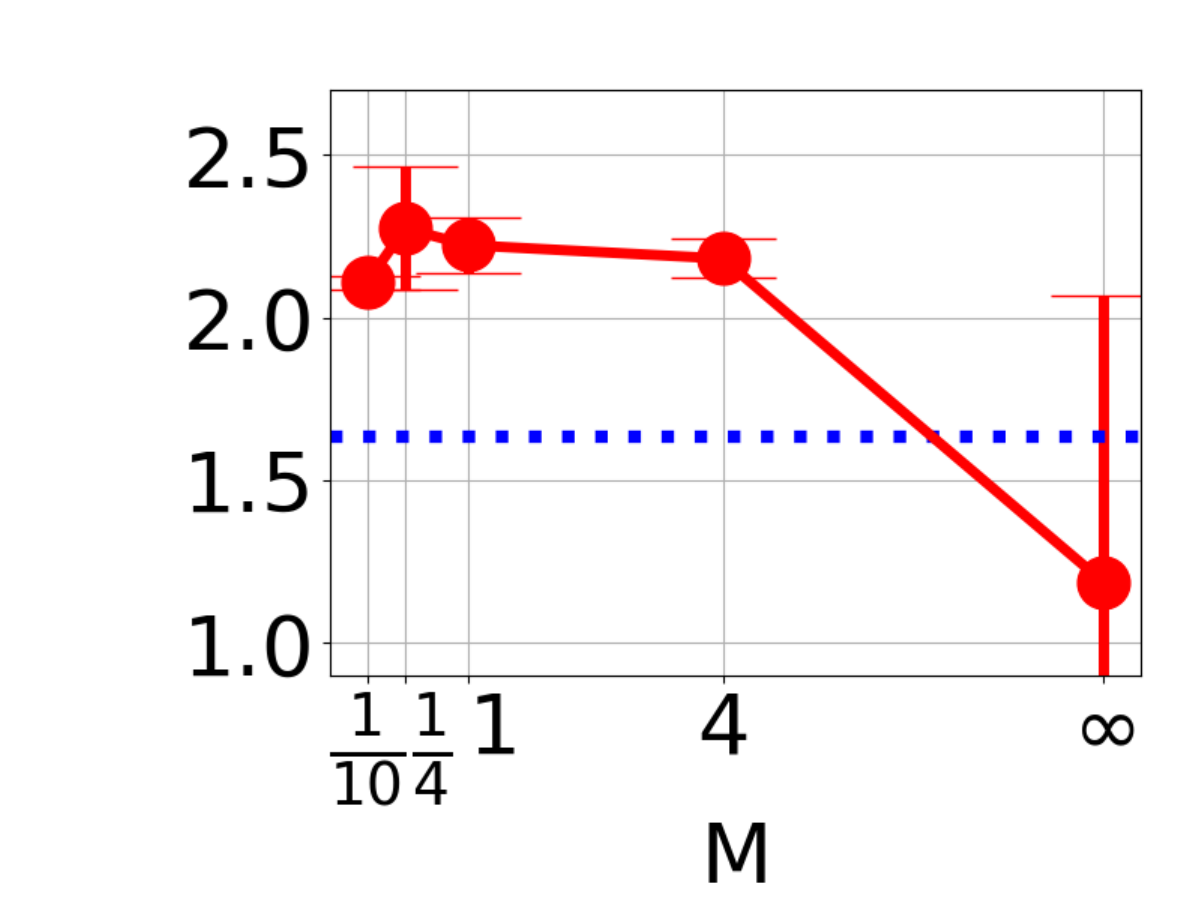} \\
\hspace{-2ex}(a) HC-k1-e & \hspace{-3ex}(b) HC-k1-m1 & \hspace{-3ex}(c) HC-k1-m2 &  \hspace{-2ex}(d) HC-k1-e-m1 & \hspace{-2ex}(e) HC-k1-e-m2 & \hspace{-2ex}(f) HC-k1-m1-m2
\end{tabular}
\caption{Ablation study on the clipping value, $M$: Performance of FACMAC+B3C with respect to $M$. $M=\infty$ (no clip) corresponds to FACMAC+BC. The performance of OMIGA is also included as a blue dotted line. Error bars represent one standard deviation.}
\label{fig:result_clipvalue}
\end{figure*}

\begin{figure*}[t!]
\begin{tabular}{c}
\includegraphics[width=0.99\linewidth]{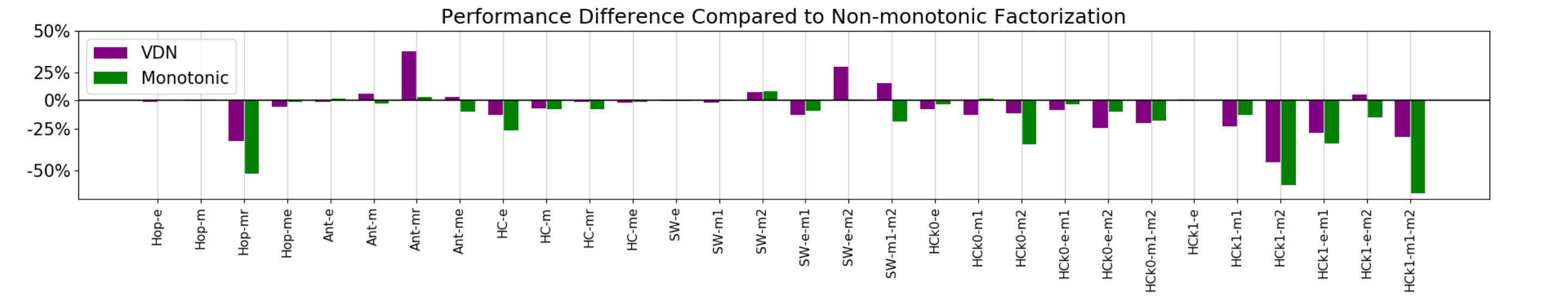} 
\end{tabular}
\vspace{-3ex}
\caption{Ablation study on value factorization: Performance differences of VDN and monotonic factorization compared to non-monotonic factorization.}
\label{fig:valuefactorization}
\end{figure*}

\textbf{Critic clipping improves performance and robustness} by enabling a higher weight on the RL objective, overcoming over-regularization while ensuring stability. As discussed in Sec. \ref{sec:overreg}, prior works employing BC regularization-based approaches suffered from over-regularization, which constrained performance to the quality of the data. This issue is inevitable without the proposed critic clipping, as increasing the weight on the RL objective induces instability, making divergence more likely, as observed in the worst-performing comparison between BC and B3C shown in Fig. \ref{fig:b3c_bc_min_difference_mpe}. However, \texttt{CC} allows for reduced BC regularization, i.e., a higher weight on the RL objective, resulting in improved performance. We investigate this by comparing FACMAC+BC and FACMAC+B3C with respect to the RL objective weight $\alpha$, while fixing $\beta=1$. The corresponding results are shown in Fig. \ref{fig:perfor_alpha}. In most cases, FACMAC+BC performs worse when $\alpha>1$, whereas FACMAC+B3C performs better or equal when $\alpha=4$ in most cases. Additionally, although FACMAC+B3C experiences a performance drop when $\alpha = 8$, it remains more stable than FACMAC+BC, demonstrating the robustness of B3C across different values of $\alpha$. Additionally, we provide the averaged performance difference between BC and B3C in multi-agent Mujoco in the Appendix~\ref{appendix:b3c_bc_mujoco}.

\subsubsection{Critic Clipping: Clipping Value}\label{sec:clipping_value}

\texttt{CC} clips the Q-function in the target value by the scaled maximum return in the dataset. This introduces a hyperparameter---the clipping value, $M$ in Eq. \ref{eq:loss_cc}. As discussed in Sec. \ref{sec:cc}, $M=1$ is sufficient since the maximum return in the dataset is already high enough to handle overestimation. We conduct an ablation study regarding the clipping value. Fig. \ref{fig:result_clipvalue} shows the performance of FACMAC+B3C by varying $M$ across $1/10$, $1/4$, $1$, and $4$. Additionally, we include the performance of FACMAC+BC, which does not use critic clipping, as $M=\infty$, and OMIGA~\cite{wang2024offline} as a blue dotted line. It is seen that $M=1$ achieves the best performance in all cases except for the HC-k1-e-m2 environment. While extremely small or large values of $M$ perform worse than $M=1$, FACMAC+B3C with any of the tested $M$ values still outperforms both FACMAC+BC and the SOTA algorithm, OMIGA. In the HC-k1-e-m2 environment, where FACMAC+B3C performs suboptimally compared to the dataset quality, reducing $M$ improves performance. Overall, $M=1$ is a reasonable choice, providing the best performance while minimizing the burden of hyperparameter tuning.

\subsubsection{Value factorization}\label{sec:ana_vf}

We here provide an ablation study of the empirical observation described in Sec.~\ref{sec:b3c_vf} that non-monotonic factorization performs better than monotonic factorization in offline settings. Fig.~\ref{fig:valuefactorization} shows the performance differences of VDN and monotonic factorization compared to non-monotonic factorization across all environments. Overall, \texttt{non-mono} consistently outperforms both \texttt{mono} and \texttt{vdn}, though the linear factorization occasionally achieves comparable or slightly higher scores in certain easy datasets.

This result contrasts with prior findings in online settings~\cite{peng2021facmac}, where monotonic factorization often yielded stable and competitive performance. We attribute this difference to the distributional shift inherent in offline learning: enforcing monotonicity restricts the representational capacity of the critic, making it difficult to approximate complex joint action dependencies when unseen samples dominate the dataset. In contrast, non-monotonic factorization allows the critic to flexibly capture nonlinear inter-agent correlations, leading to more accurate value estimation and better policy learning.


\section{Conclusion}

In this paper, we have proposed a simple regularization technique named behavior cloning with critic clipping for offline multi-agent RL that addresses overestimation and over-regularization while ensuring stability. In addition, we have investigated existing value factorization techniques in offline settings, providing the observation that non-monotonic factorization performs better than monotonic factorization. Numerical results show that B3C with non-monotonic factorization yields outstanding performance across the considered environments with diverse dataset levels. We have also provided several analyses and ablation studies to illustrate how the proposed method operates and affects learning. 

\textbf{Limitation}~~Our work lacks theoretical analysis and is empirically driven. We consider this adequate for our minimalist objective and leave theoretical extensions for future work.



\begin{acks}
This work was supported by the DARPA EMHAT Program under Agreement No. HR00112490409 and by the ONR Award No. N00014-23-1-2840.
\end{acks}



\bibliographystyle{ACM-Reference-Format} 
\bibliography{sample}


\newpage

\appendix

\section{Training Details}\label{appendix:hyper}

We implemented FACMAC+B3C based on the public code of OMIGA~\cite{wang2024offline}, and conducted the experiments on a server with dual AMD EPYC 7713 CPUs and an NVIDIA RTX 6000 Ada Generation GPU. Each experiement took about 4 hours. Code and experiment scripts are available at: \url{https://github.com/wjkim1202/B3C_OfflineMARL}

All policy and critic networks are implemented as multi-layer perceptrons (MLPs) with hidden layer sizes of 256. We use the Adam optimizer with a learning rate of $5 \times 10^{-4}$, a discount factor $\gamma = 0.99$, and Polyak target updates with $\tau = 0.005$. Gradients are clipped to a maximum norm of 1.0, and training is performed with a batch size of 128.

We now provide the key hyperparameters used for the proposed method. We have three key hyperparameters: the RL coefficient $\alpha$, the BC coefficient $\beta$, and the clipping value $M$. As discussed in the main paper, we set $M = 1$ except for one task (Ant, medium-replay dataset). For multi-agent particle environments, the values for $\alpha$ and $\beta$ are provided in Table \ref{tab:hyper_particle}. We use $M = 1$ for all cases. Next, for multi-agent Mujoco environments, we fix $\beta=1$. The values for $\alpha$ and $M$ are provided in Table. \ref{tab:hyper_mujoco}.

\begin{table}[h]
    \centering
    \caption{Hyperparameters for Various Environments}
    \begin{tabular}{lccc}
        \hline
        Environment & Dataset & $\alpha$ & $M$ \\
        \hline
        Hopper      & expert          & 16  & 1   \\
                    & medium          & 8   & 1   \\
                    & medium-replay   & 0.25 & 0.1 \\
                    & medium-expert   & 1   & 1   \\
        \hline
        Ant         & expert          & 1   & 1   \\
                    & medium          & 1   & 1   \\
                    & medium-replay   & 1   & 1   \\
                    & medium-expert   & 0.5 & 1   \\
        \hline
        HalfCheetah & expert          & 16  & 1   \\
                    & medium          & 16  & 1   \\
                    & medium-replay   & 16  & 1   \\
                    & medium-expert   & 16  & 1   \\
        \hline
        \hline
        HC\_obsk0   & expert              & 1   & 1   \\
                    & medium1             & 1   & 1   \\
                    & medium2             & 4   & 1   \\
                    & expert-medium1      & 1   & 1   \\
                    & expert-medium2      & 4   & 1   \\
                    & medium1-medium2     & 4   & 1   \\
        \hline
        HC\_obsk1   & expert              & 1   & 1   \\
                    & medium1             & 1   & 1   \\
                    & medium2             & 4   & 1   \\
                    & expert-medium1      & 4   & 1   \\
                    & expert-medium2      & 1   & 1   \\
                    & medium1-medium2     & 4   & 1   \\
        \hline
        Swimmer     & expert              & 1   & 1   \\
                    & medium1             & 1   & 1   \\
                    & medium2             & 1   & 1   \\
                    & expert-medium1      & 4   & 1   \\
                    & expert-medium2      & 1   & 1   \\
                    & medium1-medium2     & 0.5 & 1   \\
        \hline
    \end{tabular}
    \label{tab:hyper_mujoco}
\end{table}

\begin{table*}[t]
    \centering
    \caption{Hyperparameters in multi-agent particle environments}
    \begin{minipage}{0.32\linewidth}
        \centering
        \begin{tabular}{lcc}
            \hline
            CN & $\alpha$ & $\beta$ \\
            \hline
            expert-s0         & 8  & 1     \\
            expert-s1         & 8  & 1     \\
            expert-s2         & 1  & 1     \\
            expert-s3         & 4  & 1     \\
            expert-s4         & 1  & 1     \\
            medium-s0         & 8  & 0.001 \\
            medium-s1         & 16 & 0.001 \\
            medium-s2         & 8  & 0.001 \\
            medium-s3         & 8  & 1     \\
            medium-s4         & 4  & 0.001 \\
            medium-replay-s0  & 8  & 0.001 \\
            medium-replay-s1  & 8  & 0.001 \\
            medium-replay-s2  & 4  & 0.001 \\
            medium-replay-s3  & 32 & 0.001 \\
            medium-replay-s4  & 8  & 0.001 \\
            random-s0         & 8  & 0.001 \\
            random-s1         & 8  & 0.001 \\
            random-s2         & 16 & 0.001 \\
            random-s3         & 8  & 0.001 \\
            random-s4         & 16 & 0.001 \\
            \hline
        \end{tabular}
    \end{minipage}%
    \hfill
    \begin{minipage}{0.32\linewidth}
        \centering
        \begin{tabular}{lcc}
            \hline
            Tag & $\alpha$ & $\beta$ \\
            \hline
            expert-s0         & 8  & 1     \\
            expert-s1         & 1  & 1     \\
            expert-s2         & 1  & 1     \\
            expert-s3         & 1  & 1     \\
            expert-s4         & 4  & 1     \\
            medium-s0         & 8  & 1     \\
            medium-s1         & 8  & 1     \\
            medium-s2         & 8  & 0.001 \\
            medium-s3         & 8  & 1     \\
            medium-s4         & 8  & 1     \\
            medium-replay-s0  & 1  & 0.001 \\
            medium-replay-s1  & 1  & 1     \\
            medium-replay-s2  & 4  & 0.001 \\
            medium-replay-s3  & 4  & 0.001 \\
            medium-replay-s4  & 1  & 1     \\
            random-s0         & 16 & 0.001 \\
            random-s1         & 16 & 0.001 \\
            random-s2         & 16 & 0.001 \\
            random-s3         & 16 & 0.001 \\
            random-s4         & 16 & 0.001 \\
            \hline
        \end{tabular}
    \end{minipage}%
    \hfill
    \begin{minipage}{0.32\linewidth}
        \centering
        \begin{tabular}{lcc}
            \hline
            World & $\alpha$ & $\beta$ \\
            \hline
            expert-s0         & 4  & 1     \\
            expert-s1         & 2  & 1     \\
            expert-s2         & 4  & 1     \\
            expert-s3         & 4  & 1     \\
            expert-s4         & 2  & 1     \\
            medium-s0         & 4  & 1     \\
            medium-s1         & 4  & 1     \\
            medium-s2         & 4  & 1     \\
            medium-s3         & 4  & 1     \\
            medium-s4         & 4  & 1     \\
            medium-replay-s0  & 8  & 0.01  \\
            medium-replay-s1  & 8  & 0.01  \\
            medium-replay-s2  & 8  & 0.01  \\
            medium-replay-s3  & 8  & 0.01  \\
            medium-replay-s4  & 16 & 0.1   \\
            random-s0         & 16 & 0.001 \\
            random-s1         & 16 & 0.01  \\
            random-s2         & 16 & 0.001 \\
            random-s3         & 16 & 0.01  \\
            random-s4         & 16 & 0.001 \\
            \hline
        \end{tabular}
    \end{minipage}
    \label{tab:hyper_particle}
\end{table*}

\newpage

\section{Comparison between B3C and BC in the Multi-agent Mujoco environment.}\label{appendix:b3c_bc_mujoco}

\begin{figure*}[t]
\begin{tabular}{c}
\includegraphics[width=0.95\linewidth]{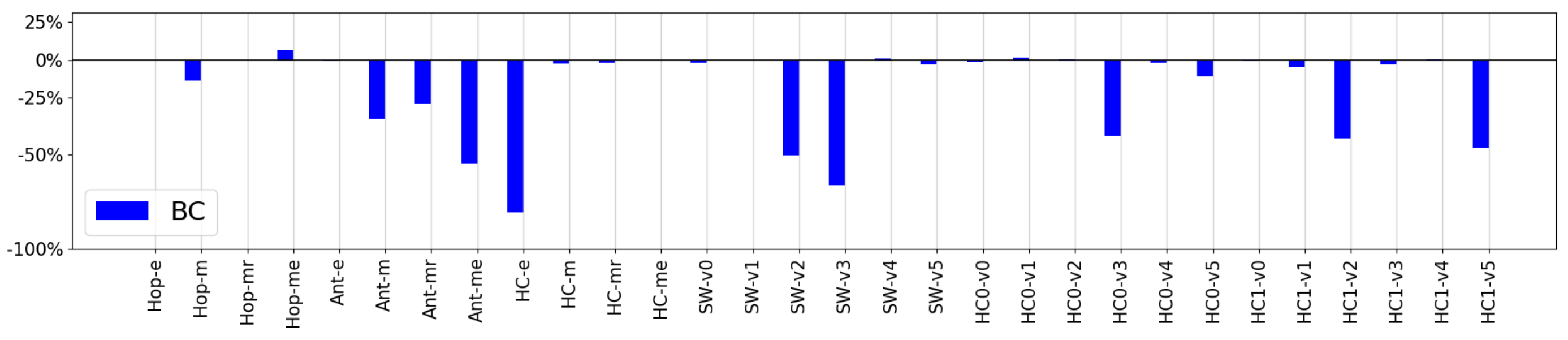} 
\end{tabular}
\caption{Performance difference FACMAC+BC compared to FACMAC+B3C, with negative values indicating that FACMAC+BC performs worse than FACMAC+B3C. A value of $-5\%$ represents that BC performs $5\%$ worse than B3C.}
\label{fig:b3c_bc_difference_mujoco}
\end{figure*}

Fig. \ref{fig:b3c_bc_difference_mujoco} shows the performance difference between FACMAC+BC and FACMAC+B3C. It is observed that B3C outperforms BC in terms of the average test return in the multi-agent Mujoco environments.

\end{document}